\RequirePackage{fix-cm}
\documentclass[smallextended]{svjour3}       
\smartqed  
\usepackage{graphicx}
\usepackage{amsmath}
\usepackage{amssymb}
\usepackage{bbm}
\usepackage{subfig}
\usepackage{color}

\allowdisplaybreaks[4]

\newtheorem{assumption}{Assumption}
\newcommand{\argmin}{\mathop{\rm arg~min}\limits}
\begin{document}

\title{Propagation Graph Estimation from Individual's Time Series of Observed States
}


\author{Tatsuya Hayashi \and Atsuyoshi Nakamura}


\institute{Tatsuya Hayashi \\
              \email{thayashi@ist.hokudai.ac.jp}           
           \\
\\
           Atsuyoshi Nakamura \\
           \email{atsu@ist.hokudai.ac.jp} \\
\\
Graduate School of Information Science and Technology, Hokkaido University, Hokkaido, Japan
}

\date{Received: date / Accepted: date}

\maketitle

\begin{abstract}
Various things propagate through the medium of individuals.
  Some individuals follow the others and take the states similar to their states a small number of time steps later.
  In this paper, we study the problem of estimating the state propagation order of individuals
  from the real-valued state sequences of all the individuals.
  We propose a method to estimate the propagation direction between individuals by the sum of the time delay of one individual's state positions from the other individual's matched state position
  averaged over the minimum cost alignments and show how to calculate it efficiently.
  The propagation order estimated by our proposed method is demonstrated to be significantly more accurate than that by a baseline method for our synthetic datasets,
  and also to be consistent with visually recognizable propagation orders for the dataset of Japanese stock price time series and biological cell firing state sequences. 
\keywords{alignment  \and time series \and propagation graph}
\end{abstract}

\section{Introduction}
Sometimes, it is very important to analyze how things such as vibration, heat, cell firing, information, virus and etc, propagated.
The objectives of such analyses are diverse from identification of the sources and the propagation routes to
learning a propagation model for prediction. Physical propagation such as vibration and heat follows physical law.
However, biological propagation such as cell firing has more ambiguous propagation rules,
and propagation through the medium of human beings such as information and virus propagation is more complex.
We study propagation analysis from the time series of the states observed at each propagating medium individual that follows ambiguous propagation rules.\\
\indent To what extent can we estimate the state propagation order
from the time series of observed states at propagating medium individuals?
Estimation of direct propagation only may be impossible because it is difficult to distinguish direct propagation from indirect propagation
and determine which individual among those with synchronous state time series has affected its states directly.  
However, we ought to be able to estimate the propagation order to some extent from the time series of individual's states only.
In fact, our proposed method can estimate the layered propagation order with accuracy of more than 70\% in our experiments using synthetic datasets generated by stochastic delay models,
where a layer is a set of individuals taking states almost synchronously. \\
\indent In this paper, we propose an alignment based method to estimate propagation direction between two individuals from their real-valued state time series.
For each pair of individuals $(i,j)$, we calculate the time delay sum of individual $j$'s states from individual $i$'s matched states averaged over all the minimum cost alignments between their state time series. Then, propagation direction between $i$ and $j$ is estimated as $i\rightarrow j$ if such averaged time delay sum is positive,
and as $j\rightarrow i$ if it is negative.
From individual pairs $(i,j)$ with non-zero average time delay sum, we construct an estimated \emph{propagation graph} whose vertices are individuals and whose edges are estimated direct propagation.
In the construction, in order to exclude indirect propagation edges, we greedily remove the edge $(i,j)$ with the largest average time delay sum if there is an indirect path from $i$ to $j$ and the delay is at least an estimated upper bound of direct propagation $\theta$,
and remove all the edges between vertices in the same estimated layer. \\
\indent According to our experiments using real-valued and symbolic time series synthetic datasets generated by stochastic delay models,
the edge sets of propagation graphs estimated by our method achieved higher recall and \emph{layer accuracy} than those by a baseline method,
where layer accuracy is the accuracy of the estimated number of steps to be taken for propagation from the source individuals to each individual. \\
\indent In order to demonstrate practical usefulness of our method,
we applied our method to propagation analyses of stock price and biological cell firing. 
For both datasets, the propagation order estimated by our proposed method is shown to be consistent with visually recognizable propagation order. 
The propagation delay is not stable for stock price propagation, but which stocks tended to follow which stocks in a given period is interesting information and automatic visualization may be useful to investors. 
Our method is considered to be useful for analyses of such unstable propagation.

\subsubsection*{Related Work}
Examining the influence of one time series $X$ on another time series $Y$ is equivalent to be examining the causal relationship between $X$ and $Y$. 
Granger causality~\cite{Granger1969} and transfer entropy~\cite{Schreiber2000} are well-known methods for investigating the causal relationship between time series. 
Even recently, extensions and applications of these methods have been energetically investigated~\cite{Quinn2015,He2017,Schwab2019}. 
In general, these methods assume that a time series is stationary. 
In Granger causality, the results also depend on a parameter in the regression model, the number of past values used. 
Usually, this parameter is selected by using an information criterion such as the Akaike Information Criterion or the Schwartz Information Criterion. 
The calculation of transfer entropy requires a sufficiently long time series to estimate probabilities. However, it is not always possible to guarantee the stationarity of a time series or to measure for a long time in real data.

In this paper, we propose a method that focuses on the delay time between time series. 
To deal with the arbitrary-time-lag influence between time series, a method integrating Granger causality and DTW was proposed~\cite{Amornbunchornvej2019}. 
This method is a generalization of Granger causality and differs from our method because it does not estimate a delay time between time series. 
Time delay estimation among signals \cite{So2008,Quazi1981} has been studied well for source localization, in which constant delay for a moment is assumed. 
Therefore, we estimate the sum of variable delays between a pair of time series for a period of some length, which gives a new perspective on causal inference between time series.

Furthermore, we also propose the way of constructing a causal graph to examine whether the propagation of effects is appropriate across individuals. 
There are many studies on information or influence propagation on networks such as studies of word-of-mouth marketing \cite{Domingos2001,Goldenberg2001TalkOT,Wang2019,Zhang2019},
epidemics \cite{Hethcote2000,Stegehuis2016,Kabir2019}, innovation diffusion \cite{UBHD2028615,Wu2016} and so on.
In most of these studies, networks are assumed to be given and not needed to be estimated though there are studies on propagation probability estimation through edges in a given network \cite{Goyal2010,Saito2008,Goyal2011,Mathioudakis2011,Varshney2017}. Recent popular studies deal with propagation through social networks \cite{Bonchi2011,Bourigault2016,Mahdizadehaghdam2016}, in which relation between users are visible and not needed to estimate in most cases. 
A method to reconstruct complex network from binary time series has developed~\cite{Ma2018,Zhang2020}. This method requires the sufficient length of binary time series because it uses the maximum-likelihood estimation of the probabilities associated with presence or absence of links.
In this paper, we also discuss how to construct a graph that shows the propagation relationship of effects among individuals.

\section{Problem Setting}

Let $I$ denote a set of individuals $\{1,\dots,N\}$.
Note that we let $[n]$ denote $\{1,\dots,n\}$ for any positive integer $n$, so $I$ is written as $[N]$.
At each time step $t=1,\dots,T$, each individual $i\in I$ takes state $s_i[t]\in Y$.
Let $s_i$ denote the string of length $T$ whose $t$th letter is $s_i[t]$, that is, $s_i=s_i[1]\cdots s_i[T]$.
We call $s_i$ as the \emph{state sequence of individual $i$}.
We consider the following state propagation between individuals.
Assume that there exists a source individual and the states propagate from individuals to individuals at each time.
As for state propagation, we assume the following.

\begin{assumption}\label{assumption:propagation}
  Each individual $i$ but the source individual, follows some other individuals $j$, and the follower $i$ takes state $s_i[t]$ similar to state $s_j[t-\Delta_{i,j}[t]]$ with small time step delay $\Delta_{i,j}[t]$ at each time step $t$.
  \end{assumption}

The state propagation can be represented by a \emph{state propagation graph} $G(V,E)$ with vertex set $V=I$ and directed edge set $E=V\times V$, in which directed edge $(i,j)\in E$ exists if and only if individual $j$ directly follows $i$.

The problem we try to solve in this paper is formalized as follows.

\begin{problem}
Given a set $\{s_1,\dots,s_N\}$ of the state sequences of individuals in $I=\{1,\dots,N\}$, estimate the state propagation graph with vertex set $I$ under Assumption~\ref{assumption:propagation}. 
\end{problem}

Note that, considering that $V$ is fixed to $I$, a solution of the above problem is estimation $\hat{E}$ of the set $E$ of directed edges.

\section{Proposed Method}

\subsection{Alignment-Based Direction Estimation}\label{alignment}

Given two state sequences $s_i, s_j$ of individuals $i, j$, how can we guess the direction of state propagation?
According to Assumption~\ref{assumption:propagation}, the propagated individual takes the state similar to that of the propagating individual small time steps later.
We propose an alignment-based method that can detect such state time delay of one individual compared to the other individual.

Let $\Pi$ denote the \emph{pair shift function set} which is defined as the set of strictly increasing function pairs $(\pi_1,\pi_2)$ from $[T]$ to $[2T-1]$ for which $\pi_1([T])\cup\pi_2([T])$ is a set of contiguous natural numbers starting from $1$, that is,
\[
\Pi=\{(\pi_1,\pi_2)\mid \begin{aligned}[t]&\pi_i(1)<\cdots<\pi_i(T) \ \ (i=1,2), \\
  &\pi_1([T])\cup\pi_2([T])=[\max(\pi_1(T)\cup\pi_2([T])]\}. \end{aligned}
\]
We let $\Pi_1$ denote the set $\Pi$ constrained by the condition that $\pi_1(1)=\pi_2(1)=1$.
An \emph{alignment} between $s_i$ and $s_j$ defined by a shift function pair $(\pi_1, \pi_2)\in \Pi$ is the state correspondence in which
$s_i[\pi_1^{-1}(k)]$ corresponds to $s_j[\pi_2^{-1}(k)]$ for $k\in \pi_1([T])\cap\pi_2([T])$, where $\pi^{-1}_i$ is the inverse function of $\pi_i$. 


There are mainly two types of alignment cost functions, warping-based and gap-based.
Consider cost function between states $w:(Y\cup \{\textvisiblespace\})\times (Y\cup \{\textvisiblespace\})\rightarrow \mathbb{R}$, where $\textvisiblespace$ is the special state corresponding to a gap.
Then, \emph{alignment cost} $S(s_i,s_j,(\pi_1,\pi_2))$is defined by
\[
S(s_i,s_j,(\pi_1,\pi_2))=\sum_{k\in \pi_1([T])\cup \pi_2([T])}w(s'_i[k],s'_j[k]),
\]
where, for $(m,\ell)=(i,1),(j,2)$,
\[
s'_m[k]=s_m[\max\{h\mid\pi_\ell(h)\leq k\}]] 
\]
in warping-based cost, and
\[
s'_m[k]=
\begin{cases}
s_m[\pi_{\ell}^{-1}(k)] & (k\in \pi_m([T]))\\
\textvisiblespace & (\text{otherwise}),
\end{cases}
\]
in gap-based cost.
Then, the \emph{minimum alignment cost} between $s_i$ and $s_j$ is $\min_{(\pi_1,\pi_2)\in \Pi_1}S(s_i,s_j,(\pi_1,\pi_2))$ for warping-based cost and $\min_{(\pi_1,\pi_2)\in \Pi}S(s_i,s_j,(\pi_1,\pi_2))$ for gap-based cost.

Time delay sum of $s_j$ from $s_i$ by the alignment between $s_i$ and $s_j$ using $(\pi_1, \pi_2)$ is defined as
\[
\sum_{k\in \pi_1([T])\cap \pi_2([T])}(\pi_2^{-1}(k)-\pi_1^{-1}(k)).
\]

We estimate the direction of state propagation between individuals $i$ and $j$ using the following rule (E).
\begin{description}
\item[(E)] The propagation direction is estimated as $i\rightarrow j$
if the time delay sum of $s_j$ from $s_i$ averaged over the minimum cost alignments between $s_i$ and $s_j$ is positive, 
and $j\rightarrow i$ if that is negative.
\end{description}

\begin{example}\label{ex:dtw}
Let $Y= \mathbb{R}$ and consider state sequences $(s_i[1],\dots,s_i[10])=(1,1,0,$\\$-1,-1,1,1,2,0,$ $-1)$ and $(s_j[1],\dots,s_j[10])=(0,1,1,0,-1,1,1,1,2,0)$ of individuals $i$ and $j$, respectively.
Define cost function $w$ as $w(x,y)=|x-y|$ and
shift functions $\pi_1$ and $\pi_2$ as $(\pi_1(1),\dots,\pi_1(10))=(1,3,4,5,6,7,9,10,11,12)$ and $(\pi_2(1),\dots,\pi_2(10))=(1,2,3,4,5,7,8,9,10,11)$.
Then, the alignment between $s_i$ and $s_j$ defined by the shift function pairs $(\pi_1,\pi_2)$ is one of minimum cost alignment with cost $2$. (See the following table.)
\begin{center}
\begin{tabular}{c|rrrrrrrrrrrr}
  $k$ & 1 & 2 & 3 & 4 & 5 & 6 & 7 & 8 & 9 & 10 & 11 & 12\\
  \hline
  $s_i[\pi_1^{-1}(k)]$ & 1 & & 1 & 0 & $-1$ & $-1$ & 1 & & 1 & 2 & 0 & $-1$\\
  $s_j[\pi_2^{-1}(k)]$ & 0 & 1 & 1 & 0 & $-1$ && 1 & 1 & 1 & 2 & 0 & \\
  $s'_i[k]$& 1 & 1 & 1 & 0 & $-1$ & $-1$ & 1 & 1 & 1 & 2 & 0 & $-1$\\
  $s'_j[k]$ & 0 & 1 & 1 & 0 & $-1$ & $-1$ & 1 & 1 & 1 & 2 & 0 & 0 \\
  $w(s'_i[k],s'_j[k])$ & 1 & 0 & 0 & 0 & 0 & 0 & 0 & 0 & 0 & 0 & 0 & 1\\
  $\pi_1^{-1}(k)$ & 1 & & 2 & 3 & 4 & 5 & 6 & & 7 & 8 & 9 & 10  \\
  $\pi_2^{-1}(k)$ & 1 & 2 & 3 & 4 & 5 & & 6 & 7 & 8 & 9 & 10 & \\  
  $\pi_2^{-1}(k)-\pi_1^{-1}(k)$ & 0 & & 1 & 1 & 1 & & 0 & &  1 & 1 & 1 &\\
\end{tabular}
\end{center}

The time delay sum of $s_j$ from $s_i$ by this alignment is
$\sum_{k\in \{1,3,4,5,7,9,10,11\}}(\pi_j^{-1}[k]-\pi_i^{-1}[k])=6$.
Considering following variations,
  \begin{center}
    \includegraphics[width=0.9\textwidth]{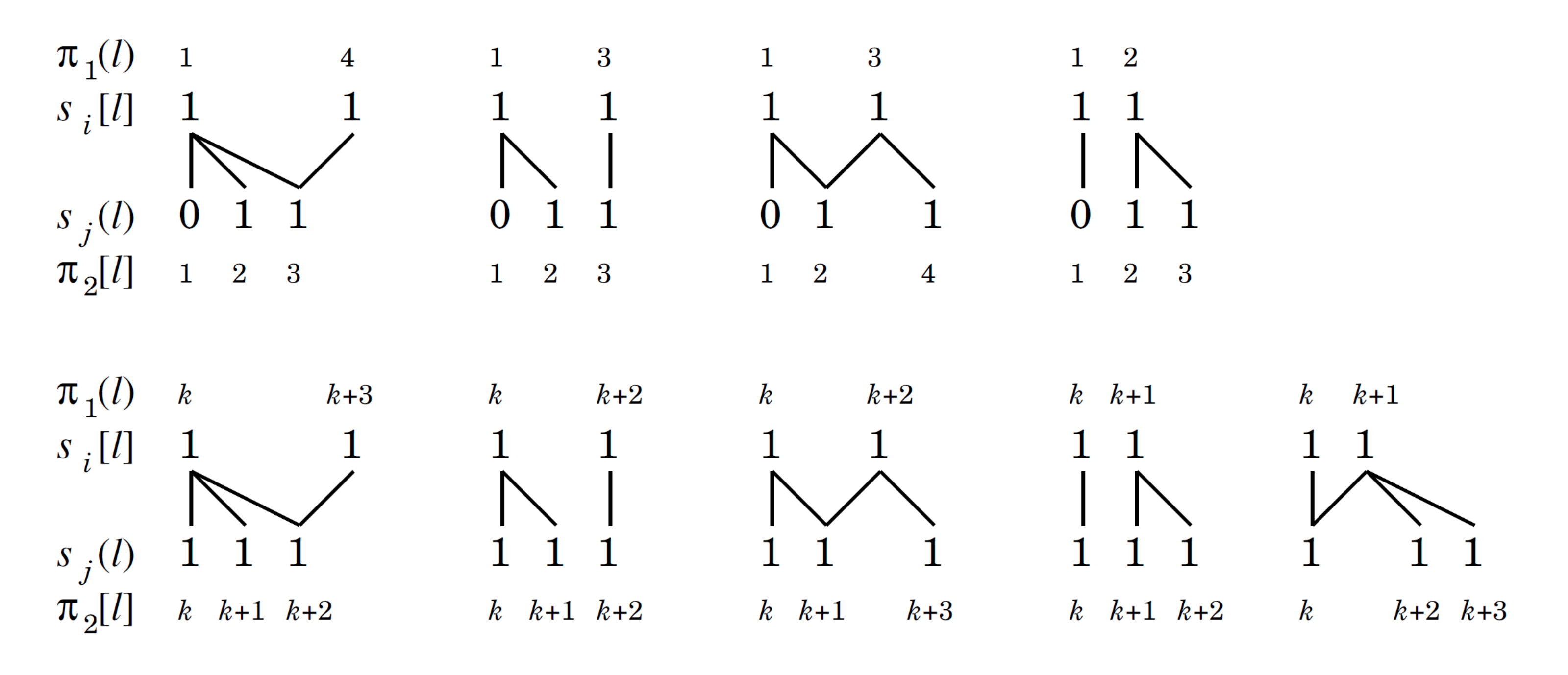}\\
\end{center}
there are 20 alignments that achieve the minimum cost $2$ and their time delay sums are 4 for 12 alignments, 5 for 7 alignments and 6 for 1 alignment.
Thus, the average time delay sum is $4.45$ and state propagation direction is estimated as $i\rightarrow j$.
\end{example}

\begin{example}\label{ex:simple}
Let $Y=\{0,1\}$ and consider state sequences $s_i=001000100$ and $s_j=000100010$.
For $\alpha\geq 2$, we consider alignments of strings $s_i$ and $s_j$ using symmetric cost function $w(x,y)$ defined as follows:
\begin{equation}
  w(x,y) = \begin{cases}
    0 & ((x,y)=(0,0),(1,1))\\
    1 & ((x,y)=(0,\textvisiblespace),(\textvisiblespace,0))\\
    \alpha & ((x,y)=(0,1),(1,0))\\
        \infty & ((x,y)=(1,\textvisiblespace),(\textvisiblespace,1)(\textvisiblespace,\textvisiblespace)).
    \end{cases}\label{costfunction}
\end{equation}
In the alignment using this cost function, each $1$-state in one sequence is strongly preferred to be aligned to $1$-state in the other sequence by shifting positions unless their position difference is large ($2\times \text{(position difference)}>\alpha$) or the number of $1$-states is different.

Consider the case with $\alpha=3$.
Then, the minimum gap-based alignment cost is $2$ and there are $6$ alignments whose alignment costs are the minimum.
One of the minimum cost alignments between $s_i$ and $s_j$ is defined by shift functions $(\pi_1(1),\dots,\pi_1(9))=(2,3,4,5,6,7,8,9,10)$ and $(\pi_2(1),\dots,\pi_2(9))=(1,2,3,4,5,6,7,8,10)$. (See the following table.)
\begin{center}
\begin{tabular}{c|cccccccccc}
  $k$ & 1 & 2 & 3 & 4 & 5 & 6 & 7 & 8 & 9 & 10\\
  \hline
  $s_i[\pi_1^{-1}(k)]$ &  &0 & 0 & 1 & 0 & 0 & 0 & 1& 0 & 0 \\
  $s_j[\pi_2^{-1}(k)]$ & 0 & 0 & 0 & 1 & 0 &0& 0 & 1 &  & 0 \\
  $s'_i[k]$& \textvisiblespace &0 & 0 & 1 & 0 & 0 & 0 & 1& 0 & 0 \\
  $s'_j[k]$ & 0 & 0 & 0 & 1 & 0 &0& 0 & 1 & \textvisiblespace & 0 \\
  $w(s'_i[k],s'_j[k])$ & 1 & 0 & 0 & 0 & 0 & 0 & 0 & 0 & 1 & 0 \\
  $\pi_1^{-1}(k)$ &  & 1& 2 & 3 & 4 & 5 & 6 & 7 & 8 & 9   \\
  $\pi_2^{-1}(k)$ & 1 & 2 & 3 & 4 & 5 & 6 & 7 & 8 & & 9  \\  
  $\pi_2^{-1}(k)-\pi_1^{-1}(k)$ &  & 1& 1 & 1 & 1 &1 & 1 & 1 & &0 \\
\end{tabular}
\end{center}
The time delay sum of this alignment is $7$. Similarly, the time delay sums of the other best alignments are
calculated as $5,6,6,7,8$, and the time delay sum averaged over all the 6 best alignments is $6.5$.
\end{example}

\subsection{Edge Set Estimation}\label{sec:edge-set-estimation}

By rule (E), directions are decided for all the individual pairs but those with zero average time delay sum.
If we let the estimated edge set $\hat{E}$ be the set of all $(i,j)\in I\times I$ with non-zero average time delay sum,
the following two issues arise:
\begin{description}
\item[P1] $\hat{E}$ contains many edges with small average time delay sum, which connects pairs of synchronized individuals.
\item[P2] $\hat{E}$ contains $(i,j)$ for which individual $i$'s state not directly but indirectly affects individual $j$'s state through the medium of some other individual $k$.
\end{description}

As a countermeasure for P2, that is, in order to delete indirectly affecting edges,
we define a candidate edge as an edge with average time delay sum larger than threshold $\theta$
and sort all the candidate edges by average time delay sum in descending order and greedily delete edge $(i,j)$ one by one for which an indirect path from $i$ to $j$ exists.
Threshold $\theta$ should be set to the estimated maximum average time delay sum of \emph{directly} affecting edges.
In the distribution over average time delay sum between all the individual pairs, average time delay sum between directly affecting pairs 
is considered to form the highest peak with high probability. So, we set $\theta$ to the first valley position larger than the highest
peak position in the distribution of the average time delay sum estimated by kernel density estimation. 

For P1, we try to partition $V$ into layers by classifying the synchronized individuals to the same layer, and then delete all the edges between vertices in the same layer.
For a given graph $G(V,E)$, define the $0$-layer set $V^E_0$ as the set\footnote{If there is no vertex with indegree $0$, define $V^E_0$ as the set of vertices for which the maximum average time delay sum among all the incoming edges is the smallest among those for all the vertices.
} of vertices with indegree $0$.
Define the $i$-layer set $V^E_i$ recursively as the set of vertices that do not belong to the $j$-layer set $V^E_j$ for any $j=0,1,...,i-1$
but have an incoming edge from some vertex in the $(i-1)$-layer set $V^E_{i-1}$.

Given a graph $G(V,\hat{E})$ with $V=I$ and the set $\hat{E}$ of directed edges $e$ whose direction is estimated by its average time delay sum $\mathrm{AD}(e)$, and threshold $\theta$, the whole process of edge set estimation is described as follows.
\begin{enumerate}
\item $e_1,\dots,e_m\gets$ sorted list of edges $e\in \hat{E}$ with $\mathrm{AD}(e)>\theta$ in descending order of $\mathrm{AD}(e)$.
\item For $e=e_1,\dots,e_m$, remove the edges $e=(i,j)\in \hat{E}$ if there exists an indirect path from $i$ to $j$.
\item Set $V_0^{\hat{E}}$ to the set of vertices in $V$ whose indegree is $0$.
\item Set $i$ to $1$. Repeat setting $V_i^{\hat{E}}$ to the set of vertices in $V\setminus \bigcup_{j=0}^{i-1}V_j^{\hat{E}}$ that has an incoming edge from a vertex in $V_{i-1}^{\hat{E}}$,  and then increasing $i$ by $1$ until $V\setminus \bigcup_{j=0}^{i-1}V_j^{\hat{E}}=V$.
\item Remove all the edges $(i,j) \in \hat{E}$ whose end points $i, j$ belong to the same layer $V_k^{\hat{E}}$ for some $k\in [N]$.
\end{enumerate}

\subsection{Calculation of Average Time Delay Sum}
In order to estimate the propagation direction between two individuals by Rule (E),
we have to calculate the time delay sum averaged over the minimum cost alignments of them.
This task is time consuming when there are many minimum cost alignments.
In this section, we propose a fast algorithm for this task. In this section, we explain the way of calculating the average time delay sum for the warping-based cost. See Appendix~\ref{sec:gap-based-calculation} for the way of calculating it with the the gap-based cost.

First, review the popular calculation algorithm for the minimum cost alignment using dynamic programming.
Consider the alignment for two strings $s_i=s_i[1]\cdots s_i[T]$ and $s_j=s_j[1]\cdots s_j[T]$.
Denote $D(t_i,t_j)$ be the minimum alignment cost between $s_i[1]\cdots s_i[t_i]$ and $s_j[1]\cdots s_j[t_j]$.
Then, $D(t_i,t_j)$ can be represented as the following recursive formula.
\begin{align*}
D(t_i,t_j)
  =& \begin{cases}
  w(s_i[1],s_j[1]) & (t_i=t_j=1)\\
  D(t_i,t_j-1)+w(s_i[1],s_j[t_j]) & (t_i=1, t_j>1)\\
  D(t_i-1,t_j)+w(s_i[t_i],s_j[1]) & (t_i>1, t_j=1)\\
  \min\{D(t_i-1,t_j),D(t_i,t_j-1),D(t_i-1,t_j-1)\} &\\
  \hspace*{\fill} +w(s_i[t_i],s_j[t_j]) & (t_i,t_j>1). 
  \end{cases}
  \end{align*}
$D(T,T)$ is the minimum alignment cost between $s_i$ and $s_j$, and $D(T,T)$ can be calculated
by calculating $D(t_i,t_j)$ in the order of $(t_i,t_j)=(1,1),\cdots,(1,T),$ $(2,1),\cdots,(2,T),\cdots,(T,1),\cdots,(T,T)$
using the above recursive formula.

Consider the directed graph $G=(V,E)$ with
\begin{align*}
  V=&\{(t_i,t_j)\mid t_i,t_j\in\{1,\dots,T\}\}\\
  E=&\begin{aligned}[t]
    & 
    \{((t_i,t_j-1),(t_i,t_j))\mid D(t_i,t_j)=D(t_i,t_j-1)+w(s_i[t_i],s_j[t_j])\}\\
   & 
    \cup  \{((t_i-1,t_j),(t_i,t_j))\mid D(t_i,t_j)=D(t_i-1,t_j)+w(s_i[t_i],s_j[t_j])\}\\
   & 
    \cup  \{((t_i-1,t_j-1),(t_i,t_j))\mid D(t_i,t_j)=D(t_i-1,t_j)+w(s_i[t_i],s_j[t_j])\} .
    \end{aligned}
\end{align*}
Then, all the paths from $(1,1)$ to $(T,T)$ on $G$ correspond to the minimum cost alignments.

\begin{figure}[t]
\begin{center}
  \includegraphics[scale = 0.3]{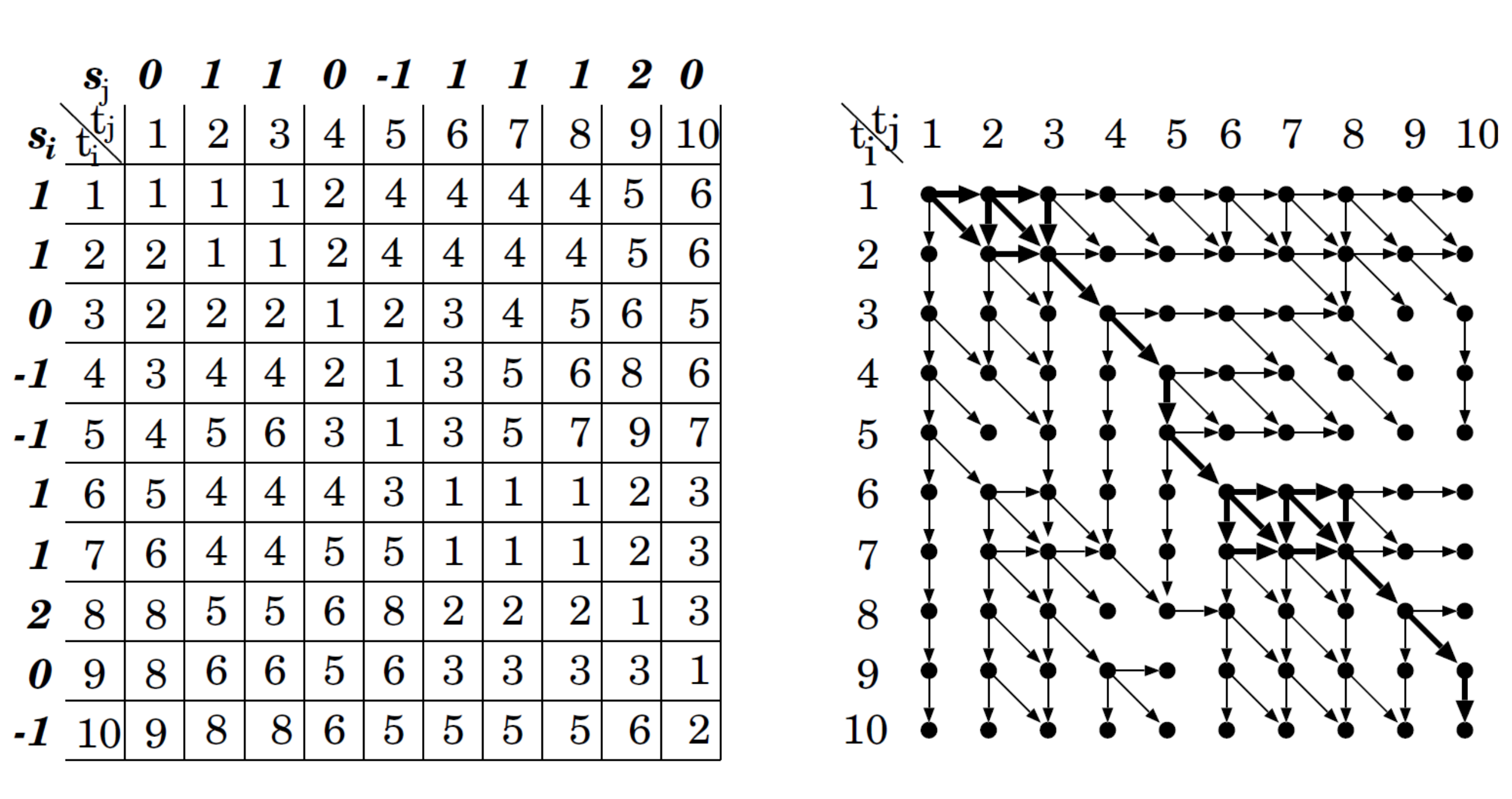}\\
  \small  $D(t_i,t_j)$\hspace*{3.8cm}$G$\\
 \includegraphics[scale = 0.3]{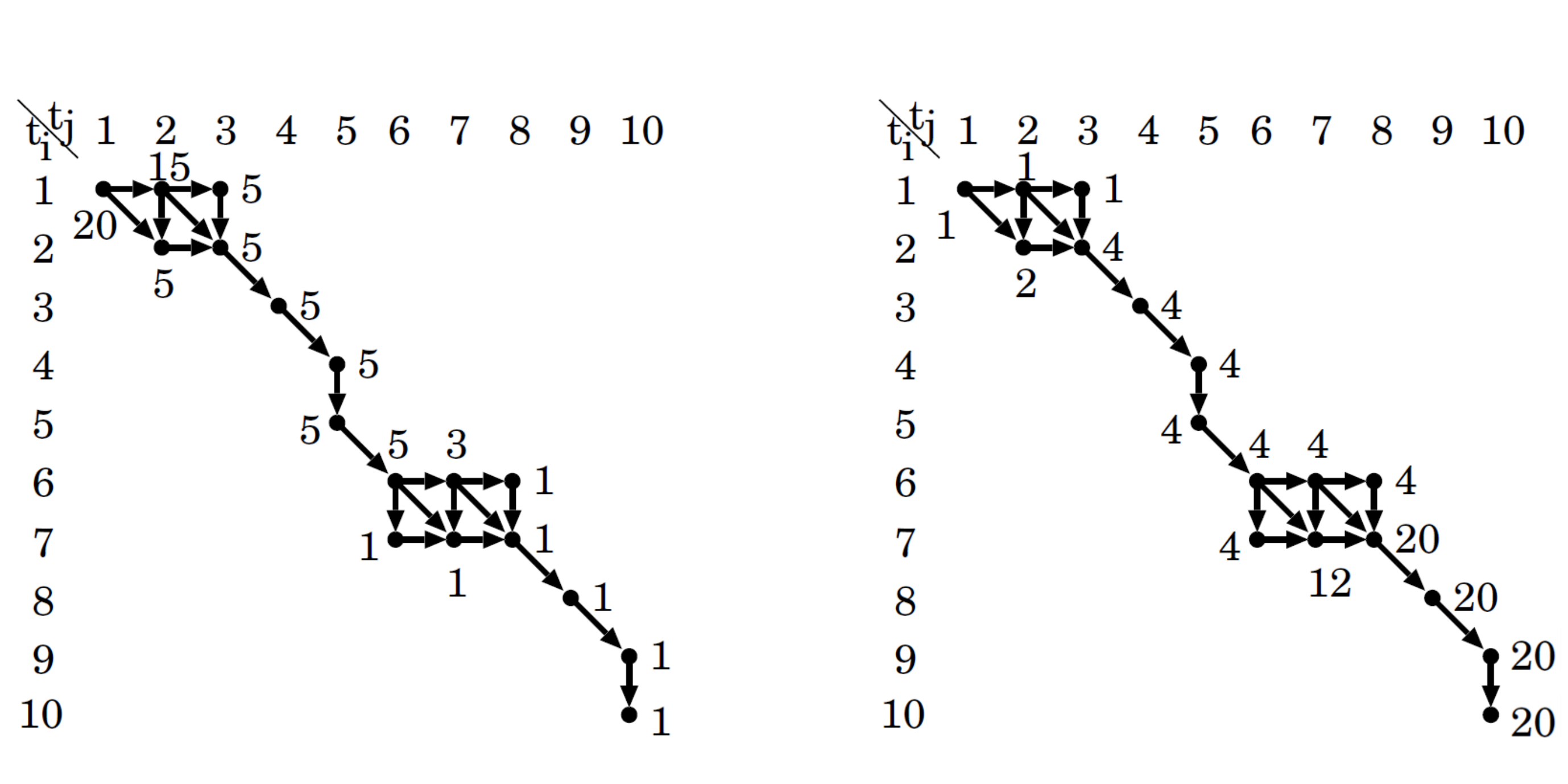}\\
     $B(t_i,t_j)$\hspace*{3.5cm}$F(t_i,t_j)$
\end{center}
  \caption{$D$ for strings $s_i$ and $s_j$ with cost function $w(x,y)=|x-y|$ in Example~\ref{ex:dtw} and its corresponding graph $G$, and $B$ and $F$ on the minimum cost paths. The directed edges in the paths from $(1,1)$ to $(10,10)$ on $G$ are bolded.
$B(t_i,t_j)$s and $F(t_i,t_j)$s for $(t_i,t_j)$ only in the paths corresponding to the minimum cost alignments are shown and $B(t_i,t_j)$s and $F(t_i,t_j)$s for other $(t_i,t_j)$ are $0$ and not needed to be calculated.
  }\label{fig:DandG}
\end{figure}

\begin{example}
  $D$ for strings $s_i$ and $s_j$ with cost function $w(x,y)=|x-y|$ in Example~\ref{ex:dtw} and its corresponding graph $G$ are shown in Fig.~\ref{fig:DandG}.
  The 20 minimum cost alignments correspond to the paths from $(1,1)$ to $(10,10)$ on $G$.
\end{example}

To calculate the time delay sum averaged over the minimum cost alignments, it is enough to calculate two values, the number of the minimum cost alignments
and the sum of time delay over matched positions in the minimum cost alignments.

The number of the minimum cost alignments between $s_i$ and $s_j$ coincides with the number of paths from $(1,1)$ to $(T,T)$ in $G$.
Let $B(t_i,t_j)$ be the number of paths from $(t_i,t_j)$ to $(T,T)$. What we want to calculate is $B(1,1)$.
$B(t_i,t_j)$ can be represented as the following recursive formula:
\begin{align*}
 B(t_i,t_j)
  =&\begin{cases}
1 & \!\!\!\!\!\!\!\!((t_i,t_j)=(T,T))\\
\mathbbm{1}\{((t_i,t_j),(t_i,t_j\!+\!1))\!\in\! E\}B(t_i,t_j\!+\!1) & (t_i\!=\!T, t_j\!<\!T)\\
\mathbbm{1}\{((t_i,t_j),(t_i\!+\!1,t_j))\!\in\! E\}B(t_i\!+\!1,t_j) & (t_i\!<\!T, t_j\!=\!T)\vspace*{1mm}\\
\mathbbm{1}\{((t_i,t_j),(t_i,t_j\!+\!1))\!\in\! E\}B(t_i,t_j\!+\!1) &\\
\ \ +\mathbbm{1}\{((t_i,t_j),(t_i\!+\!1,t_j))\!\in\! E\}B(t_i\!+\!1,t_j) &\\
\ \ +\mathbbm{1}\{((t_i,t_j),(t_i\!+\!1,t_j\!+\!1))\!\in\! E\}B(t_i\!+\!1,t_j\!+\!1) & (t_i,t_j\!<\!T),
\end{cases}
\end{align*}
where $\mathbbm{1}\{\cdot\}$ is an indicator function, that is, $\mathbbm{1}\{\cdot\}=1$ if `$\cdot$' holds and $0$ otherwise.
$B(1,1)$ can be obtained by starting from $B(T,T)$ and calculating $B(t_i,t_j)$ in reverse lexicographic order of $(t_i,t_j)$ using this recursive formula.

\begin{example}
  $B$s for strings $s_i$ and $s_j$ with cost function $w(x,y)=|x-y|$ in Example~\ref{ex:dtw} is shown in Fig.~\ref{fig:DandG}.
  The number of the minimum cost alignments can be calculated as $20$ from $G$ using the above recursive formula.
\end{example}

Finally, we explain how to efficiently calculate the sum of time delay over matched positions in the minimum cost alignments.
The pairs of the matched positions $(t_i,t_j)$ correspond to diagonal edges $((t_i-1,t_j-1),(t_i,t_j))$ in $G$.
The time delay of $s_j$ from $s_i$ for the matched position $(t_i,t_j)$ is $t_j-t_i$.
The number of the minimum cost alignments that contains matched position $(t_i,t_j)$ coincides with
the number of the paths from $(1,1)$ to $(T,T)$ in $G$ that include directed edge $((t_i-1,t_j-1),(t_i,t_j))$.
Let $F(t_i,t_j)$ be the number of paths from $(1,1)$ to $(t_i,t_j)$ in $G$ and let $E^\ast$ denote the set of directed edges in $E$
that are included in the paths corresponding to the minimum cost alignments.
Then, the sum of time delay over matched positions in the minimum cost alignments is calculated as
\[
\sum_{((t_i-1,t_j-1),(t_i,t_j))\in E^\ast} (t_j-t_i)F(t_i-1,t_j-1)B(t_i,t_j).
\]
Note that $F(t_i,t_j)$ can be also expressed by recursive formula as follows:
\begin{align*}
F(t_i,t_j)
  =&\begin{cases}
1 & \!\!\!\!\!\!\!\!((t_i,t_j)=(1,1))\\
\mathbbm{1}\{((t_i,t_j\!-\!1),(t_i,t_j))\!\in\! E\}F(t_i,t_j\!-\!1) & (t_i\!=\!1, t_j\!>1)\\
\mathbbm{1}\{((t_i-1,t_j),(t_i,t_j))\!\in\! E\}F(t_i\!-\!1,t_j) & (t_i\!>\!1, t_j\!=\!1)\vspace*{1mm}\\
\mathbbm{1}\{((t_i,t_j\!-\!1),(t_i,t_j))\!\in\! E\}F(t_i,t_j\!-\!1) &\\
\ \ +\mathbbm{1}\{((t_i\!-\!1,t_j),(t_i,t_j))\!\in\! E\}F(t_i\!-\!1,t_j) &\\
\ \ +\mathbbm{1}\{((t_i\!-\!1,t_j\!-\!1),(t_i,t_j))\!\in\! E\}F(t_i\!-\!1,t_j\!-\!1) & (t_i,t_j\!>\!1),
\end{cases}
\end{align*}

\begin{remark}
The number of the paths from $(1,1)$ to $(T,T)$ can be obtained as $F(T,T)$ using the above recursive formula for $F$ similarly as $B(1,1)$ using the recursive formula for $B$.
However, $B(1,1)$ is more appropriate than $F(T,T)$ for calculating the number of the minimum cost alignments because we can reach $(1,1)$ from $(T,T)$ by going up the directed edges in $E^\ast$ only
without knowing $E^\ast$
while the knowledge of $E^\ast$ is needed to reach $(T,T)$ from $(1,1)$ by going down the directed edges in $E^\ast$ only.
By calculating $B$ first, we can use the knowledge of $E^\ast$ to calculate $F$ for the necessary pairs $(t_i,t_j)$ only.
\end{remark}

\begin{example}
  $F$ for strings $s_i$ and $s_j$ with cost function $w(x,y)=|x-y|$ in Example~\ref{ex:dtw} is shown in Fig.~\ref{fig:DandG}.
  From the values in B and F, we can calculate the sum of time delay over matched positions in the minimum cost alignments as
 $0\cdot 1\cdot 5 +1\cdot 1\cdot 5 + 2\times 1\cdot 4\cdot 5 + 0\cdot 4\cdot 5 +0\cdot 4\cdot 1
    +1\cdot 4\cdot 1 + 2\times 1\cdot 20\cdot 1 =89$.
  Thus, the time delay sum averaged over the minimum cost alignments is $89/20=4.45$, which coincides with calculation in Example~\ref{ex:dtw}.
\end{example}

\subsection{Space and Time Complexities}

The time and space complexities to construct tables $D$, $B$ and $F$, are $O(T^2)$.
So, propagation direction estimation for a pair of individuals can be processed in time and space $O(T^2)$.
There are $O(N^2)$ pairs of cells, thus the estimation for all the pairs takes $O(N^2T^2)$ time and $O(N^2+T^2)$ space totally.
For edge set estimation, greedy removing the edges with average time delay sum larger than threshold for which indirect paths exist, takes $O(N^3)$ time and $O(N^2)$ space because sorting edges takes $O(N^2\log N)$ time and $O(N^2)$ space and indirect path existence checking takes $O(N)$ time and space per edge.
Layer partition and removing edges between the same layer vertices also take $O(N^2)$ time and space. 
Totally, our method runs in time $O(N^3+N^2T^2)$ and space $(N^2+T^2)$ in the worst case.

\section{Experiments}

In this section, we experimentally show effectiveness of our method using synthetic and real world datasets. 
The gap-based cost $w$ defined by Eq. (\ref{costfunction}) with $\alpha=3$ is used by the proposed method using gap-based cost in all the experiments for binary state propagation.

\subsection{Experiments Using Synthetic Datasets} 

First, we evaluate how accurate the estimated edge set $\hat{E}$ by the proposed method is for the real-valued and binary state sequence dataset generated from a delay model with a given ground truth propagation graphs $G(V,E)$.

\subsubsection{Ground Truth Graphs and Datasets}

\textbf{[Real-valued State Propagation]}
 We generate the dataset using ground truth propagation graph $G(V,E)$ shown on the left of Fig.~\ref{fig:ex2}.
 The length-$100$ time series $s_i[1]\cdots s_i[100]$ for $i=1,\dots,10$ are generated as following steps.
 Note that $\text{in}(i)$ denotes the set of nodes from which edges come to node $i$ and operator $\%$ is modulus operator.
\begin{description}
\item[Step 1] Generate an i.i.d. sequence $s_1[1],\dots,s_1[100]\sim N(0,5^2)$.
\item[Step 2] Set $U$ to $\{2,\dots,10\}$. 
\item[Step 3] While $U\neq\emptyset$, generate a sequence $s_i$ for $i\in U$ with $\text{in}(i)\cap U=\emptyset$ as follows: 
\begin{enumerate}
\item $s_i[1],s_i[2]\sim N(0,5^2)$, $\Delta_{j,i}[2]\gets 1$ or $2$ randomly.
\item For $t=3,4,\dots,100$, generate $s_i[t]$ as
\begin{align*}  
  \Delta_{j,i}[t]\gets &\begin{cases}\Delta_{j,i}[t-1] & \text{ with prob. } 3/4\\ \Delta_{j,i}[t-1]\%2+1 & \text{ with prob. } 1/4\end{cases}\\
    \epsilon\gets & \text{random value generated according to } N(0,1)\\
    s_i[t]\gets& \left(\sum_{j\in \text{in}(i)}s_j[t-\Delta_{j,i}[t]]\right)/|\text{in}(i)|+\epsilon.
\end{align*}
 \item $U\gets U\setminus \{i\}$
\end{enumerate}
\end{description}
We generated 100 datasets using this procedure in our experiment.\\

\noindent
\textbf{[Binary State Propagation]}
The dataset is generated by propagation model in which individuals are located in $2$-dimensional real space and state-$1$ of individual $j$
is propagated from individuals $i$ within some distance, then the ground truth graph $G_{\mathbb{B}}(V,E)$ is generated from the dataset and individuals' location information.
Note that the proposed method estimates $E$ without individuals' location information.
Given a parameter $0<p\leq 1$ of the state-$1$ propagation probability, the length-$200$ time series $s_i[1]\cdots s_i[200]$ for $i=1,\dots,50$ is generated as following steps.
\begin{description}
\item[Step 1] For $i=1,\dots,50$, the location $r_i$ of individual $i$ randomly selected according to uniform distribution over $[0,M]^2$.
\item[Step 2] Set $s_1[t]=1$ for $t \mathrm{\ mod\ } 10 = 1$ and $s_1[t]=0$ otherwise for $t=1,\dots,200$, where $\mathrm{mod}$ is modulo operation.
\item[Step 3] For $i=2,\dots,50$ and $t=1,\dots,200$, set $s_i[t]=1$ with probability $p$ if the following two conditions 
\begin{enumerate}
\item $\exists j$ s.t. $\|r_j-r_i\|\leq 35$, $s_j[t-1]=1$ (there is an individual within distance $35$ that takes state $1$ at just one step before) and
\item $s_i[t-k]=0$ for all $k=1,2,\dots,\min\{5,t-1\}$ (state-$1$ interval of each individual is at least $5$ for avoiding immediate inverse propagation),
\end{enumerate}
are satisfied and set $s_i[t]=0$ otherwise.
\end{description}

From the dataset $\{s_1,\dots,s_{50}\}$ generated above and location information $\{r_1,\dots,r_{50}\}$, edge set $E$ of the  ground truth propagation graph $G_{\mathbb{B}}(V,E)$
is created as follows.
Let $n(i,j)$ denote the number of individual $j$'s state $1$ caused by individual $i$'s state $1$, that is,
\[
n(i,j)=|\{t \in \{1,\dots,200\}\mid \begin{aligned}[t] s_i[t-1]=1, s_j[t]=1,&\\ \|r_j-r_i\|<35 & \}|,\end{aligned}
\]
where $|\cdot|$ denotes the number of elements in set `$\cdot$'.
Then, $E$ is defined as
\[
E=\{(i,j)\in V\times V\mid n(i,j)>n(j,i)\}.
\]
A ground truth graph $G_{\mathbb{B}}(G,E)$ for one dataset with $p=0.95$ is shown on the left of Fig.~\ref{fig:ex1}.


In the experiment, we generate $100$ datasets and corresponding ground truth graphs for each $p=1.00, 0.95, 0.90, 0.80, 0.70, 0.60, 0.50$. 

\subsubsection{Evaluation Measures}\label{sec:eval}

As a direct evaluation measure of delay estimations, we define \emph{mean absolute error of average time delay (MAEATD)} as
follows. For $(i,j)\in E$, define $D_{i,j}$ as $D_{i,j}=\sum_{t=a}^{T}\Delta_{i,j}[t]$ and let $\hat{D}_{i,j}$ denote its estimation, where $a$ is the maximum possible time delay in the ground truth model. Then, 
MAEATD for estimations is defined as $\text{MAEATD}=\frac{1}{|E|(T-a)}\sum_{(i,j)\in E}|\hat{D}_{i,j}-D_{i,j}|$.
Using directed edge set $E$ of the ground truth propagation graph, we evaluate an estimated directed edge set $\hat{E}$
in terms of \emph{precision (Prec)}, \emph{recall (Rec)} and \emph{$F$-measure (FM)} defined as
\begin{align*}
  \text{Prec}= \frac{|E \cap \hat{E}|}{|\hat{E}|},\ 
  \text{Rec} = \frac{|E \cap \hat{E}|}{|E|} \text{ and } 
  \text{FM} = \frac{2\,\,\text{Prec} \cdot \text{Rec}}{\text{Prec} +\text{Rec}}.
\end{align*}

It is very difficult to estimate $E$ with high precision in our setting, so we also evaluate $\hat{E}$ in terms of looser measures.
We can also consider layer partition $V_0^{\hat{E}}, V_1^{\hat{E}},\cdots$ for $G(V,\hat{E})$ like layer partition $V_0^E, V_1^E,\cdots$ that is defined in Sec.~\ref{sec:edge-set-estimation} for the ground truth propagation graph $G(V,E)$.
Then, we define \emph{layer accuracy (LA)} and \emph{Mean layer difference (MLD)} of $\hat{E}$ as
\[
\text{LA}=\frac{\sum_{i=0}^N|V^E_i\cap V^{\hat{E}}_i|}{|V|} \text{ and } \text{MLD}=\frac{\sum_{i=1}^N|\ell^E(i)-\ell^{\hat{E}}(i)|}{N},
\]
where $\ell^E(i)$ denote the individual $i$'s belonging to layer in $G(V,E)$, that is,
$\ell^E(i)=j \stackrel{\text{def}}{\Leftrightarrow} i\in V^E_j$.

As a baseline method, we consider a method using optimal constant time delay $\hat{D}_{i,j}$ of individual $j$'s state from individual $i$'s state,
which is defined as
\[
\hat{D}_{i,j}=\argmin_{-T/2<\Delta\leq T/2}\sum_{t=1}^{T}(s_j[t]-s_i[(t+(T-1)-\Delta)\%T+1])^2,
\]
where $\%$ is modulus operator. If there are multiple candidates for $\hat{D}_{i,j}$, we adopt $\hat{D}_{i,j}$ with the smallest absolute value.
Using $\hat{D}{i,j}$, propagation direction is estimated as $i\rightarrow j$ if $\hat{D}_{i,j}>0$ and $j\rightarrow i$ if $\hat{D}_{i,j}<0$.
We construct estimated edge set $\hat{E}$ of the baseline method by applying the procedure proposed in Sec.~\ref{sec:edge-set-estimation} using $\hat{D}_{i,j}$ instead of the average time delay sum of $s_j$ from $s_i$.

\subsubsection{Results}
\textbf{[Real-valued State Propagation]}
Performance comparison with the baseline method by the evaluation measures in Sec.~\ref{sec:eval} is shown below.
\begin{center}
\scalebox{0.72}{
  \begin{tabular}{ccccccc}
\hline\noalign{\smallskip}
      Method & MAEATD& Prec & Rec & FM & LA & MLD \\
\noalign{\smallskip}\hline\noalign{\smallskip}
      Baseline & 0.462 {\scriptsize ($\pm$0.004)} & 0.367 {\scriptsize ($\pm$0.005)}  & 0.431 {\scriptsize ($\pm$0.024)} & 0.390 {\scriptsize ($\pm$0.014)} & 0.402 {\scriptsize ($\pm$0.005)} & 0.662 {\scriptsize ($\pm$0.016)}  \\
      Proposed &0.317 {\scriptsize ($\pm$0.014)}  &0.509 {\scriptsize ($\pm$0.030)}  &0.621 {\scriptsize ($\pm$0.049)}  &0.556 {\scriptsize ($\pm$0.037)}  &0.772 {\scriptsize ($\pm$0.048)}  &0.275 {\scriptsize ($\pm$0.073)}  \\
\noalign{\smallskip}\hline
    \end{tabular}
}
\end{center}
Note that the values in the table are averaged over 100 datasets and the parenthesized values are their $95\%$ confidence intervals.  
You can see that our method significantly outperforms the baseline method in all the measures.
\begin{figure}[tb]
  \centering
  \begin{tabular}{c@{\ \ \ \ \ }c@{\ \ \ \ \ }c}
  \includegraphics[width=3cm]{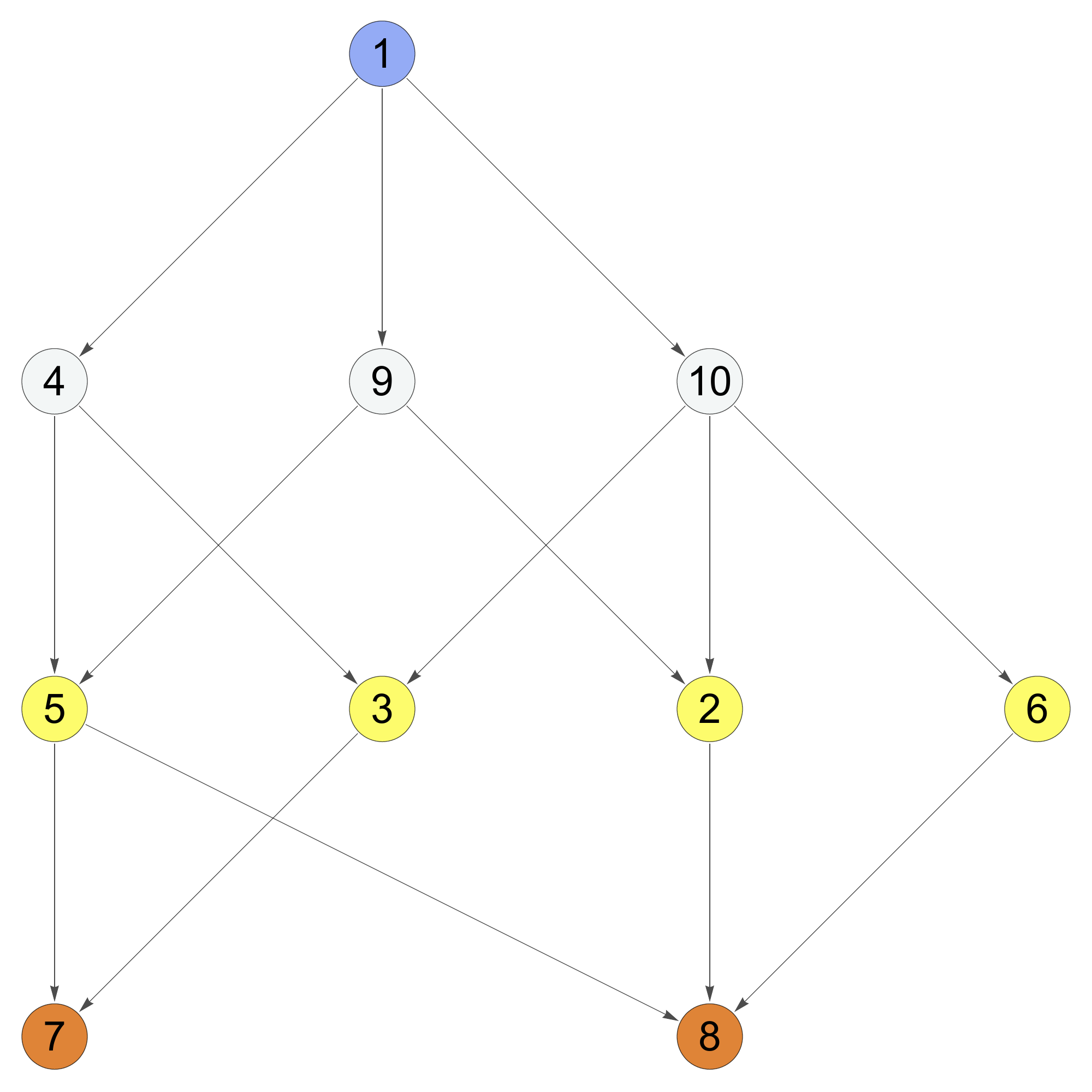}& \includegraphics[width=4cm]{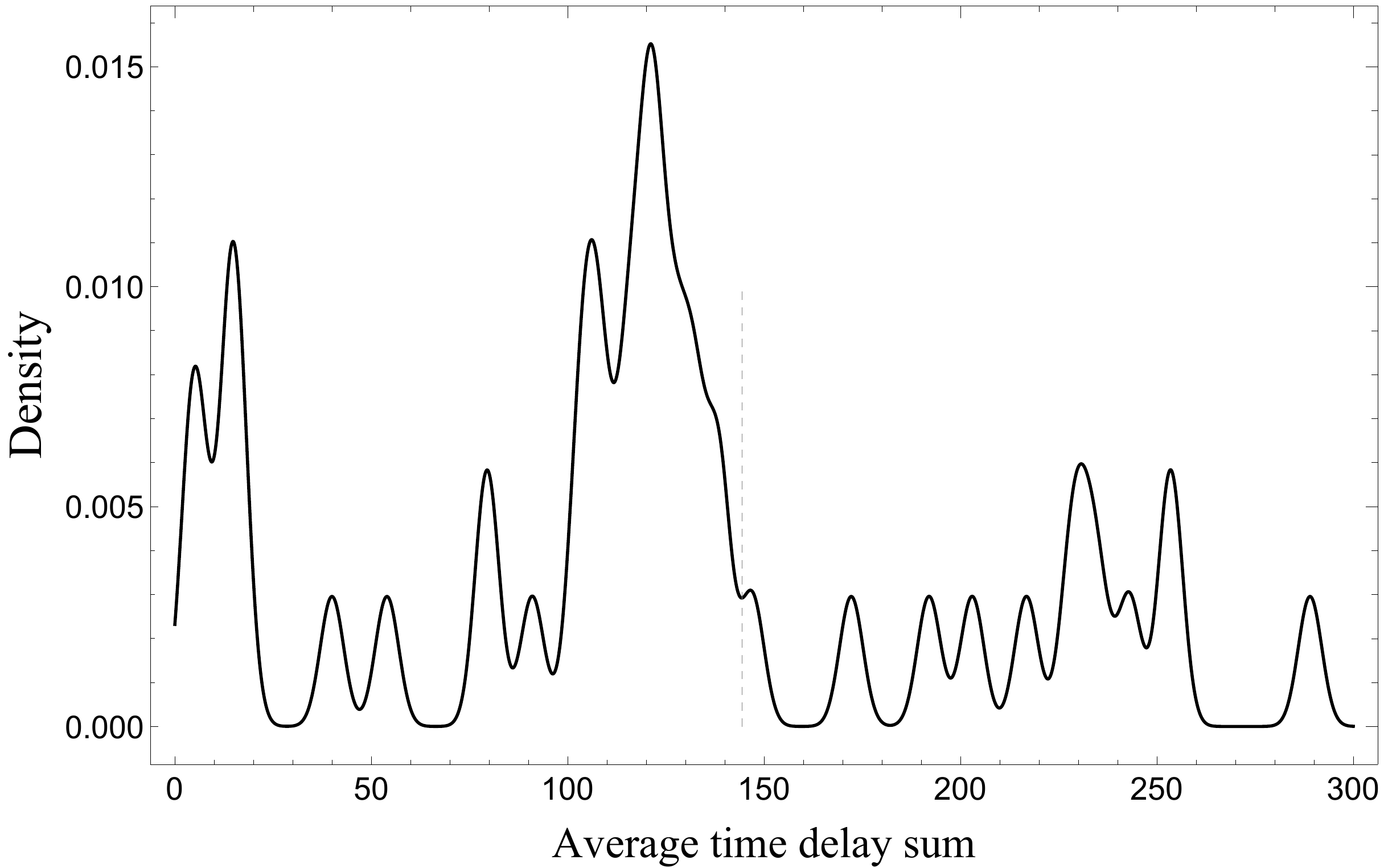} &\includegraphics[width=3cm]{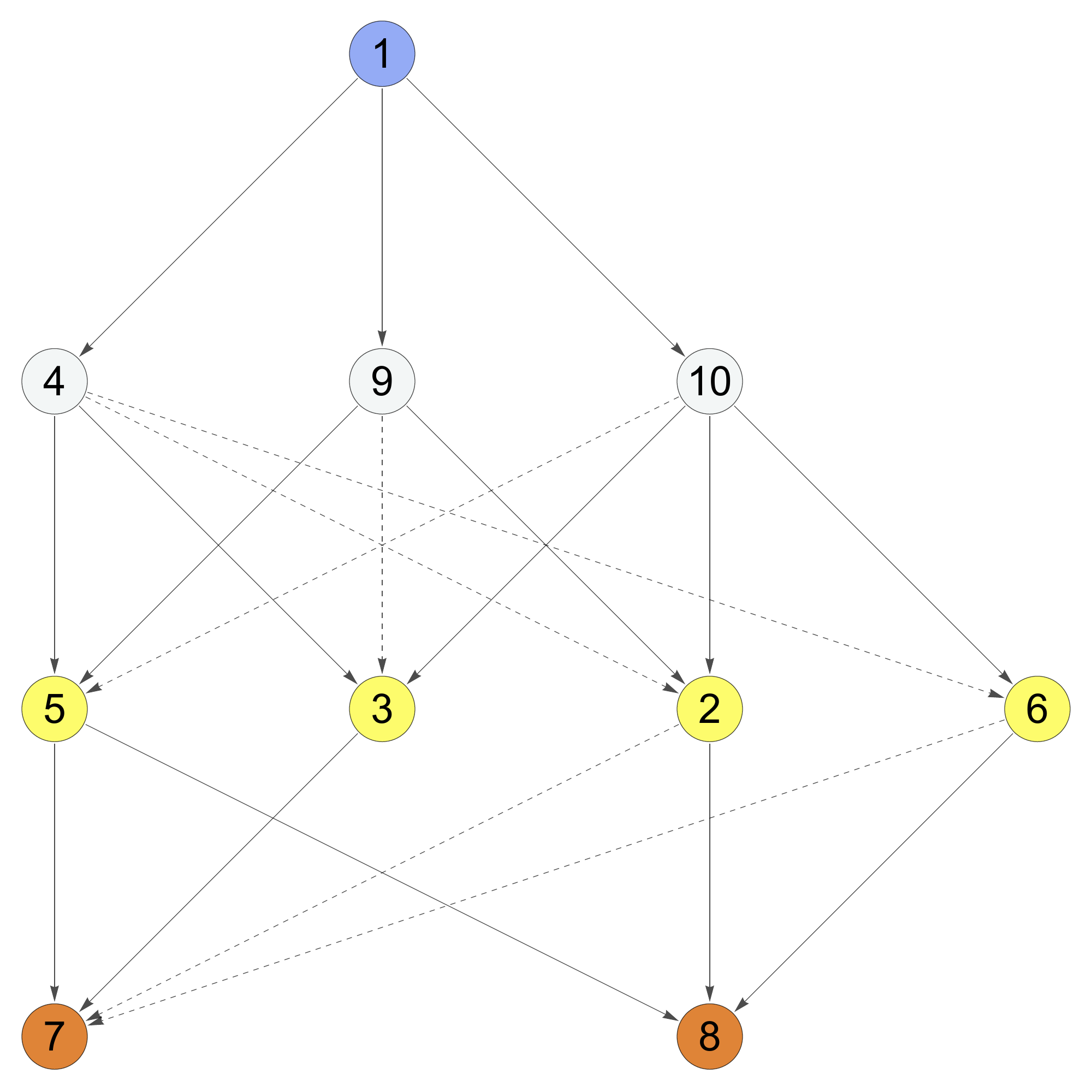}\\
  $G(V,E)$ & &$G(V,\hat{E})$\\
  \end{tabular}
   \caption{[Left] Ground truth graph $G(V,E)$. [Center] The probability density of average time delay sum estimated by kernel density estimation for one of the synthetic datasets. [Right] The estimated graph $G(V,\hat{E})$ from one of the datasets by the proposed method. 
    In $G(V,\hat{E})$, solid and dashed arrows are edges in $E$ and $\hat{E}\setminus E$, respectively.
    Each node's color indicates its belonging layer.
}\label{fig:ex2}
\end{figure}

The estimated propagation graph $G(V, \hat{E})$ by the proposed method for one of the synthetic datasets is shown in the right figure of Fig. \ref{fig:ex2}.
Parameter $\theta$ is set to $144.401$ from the estimated distribution (the center of Fig.~\ref{fig:ex2}). For this dataset, there are some falsely detected edges but all the edges in $E$ are correctly detected keeping the layer structure. \\

\noindent
\textbf{[Binary State Propagation]}
\begin{table}[tb] 
  \begin{center}
    \caption{Estimation performance of the baseline method and our proposed method using gap-based cost averaged over $100$ datasets for 6 values of parameter $p$:$0.50, 0.60, 0.70, 0.80, 0.90, 0.95, 1.00$.}\label{tab:result}
\scalebox{0.85}{
    \begin{tabular}{clccccc}
     $p$ & Method  & Prec & Rec & FM & LA & MLD\\ \hline
     1.00  &  baseline     & 0.187 {\scriptsize ($\pm$0.005)} & 0.953 {\scriptsize ($\pm$0.037)} & 0.301 {\scriptsize ($\pm$0.010)} & 0.309 {\scriptsize ($\pm$0.020)} & 1.080 {\scriptsize ($\pm$0.068)} \\
     1.00  & proposed & 0.281 {\scriptsize ($\pm$0.007)} & 1.000 {\scriptsize ($\pm$0.000)} & 0.437 {\scriptsize ($\pm$0.009)} & 1.000 {\scriptsize ($\pm$0.000)} & 0.000 {\scriptsize ($\pm$0.000)} \\\hline 
     0.95  & baseline     & 0.167 {\scriptsize ($\pm$0.006)} & 0.624 {\scriptsize ($\pm$0.040)} & 0.254 {\scriptsize ($\pm$0.012)} & 0.302 {\scriptsize ($\pm$0.021)} & 1.128 {\scriptsize ($\pm$0.069)} \\
     0.95  & proposed &   0.303 {\scriptsize ($\pm$0.010)} & 0.997 {\scriptsize ($\pm$0.006)} & 0.462 {\scriptsize ($\pm$0.011)} & 0.987 {\scriptsize ($\pm$0.021)} & 0.037 {\scriptsize ($\pm$0.060)} \\\hline
     0.90  & baseline     & 0.176 {\scriptsize ($\pm$0.007)} & 0.432 {\scriptsize ($\pm$0.027)} & 0.242 {\scriptsize ($\pm$0.011)} & 0.305 {\scriptsize ($\pm$0.020)} & 1.114 {\scriptsize ($\pm$0.065)} \\
     0.90  & proposed &   0.302 {\scriptsize ($\pm$0.010)} & 0.989 {\scriptsize ($\pm$0.018)} & 0.461 {\scriptsize ($\pm$0.012)} & 0.953 {\scriptsize ($\pm$0.041)} & 0.108 {\scriptsize ($\pm$0.097)} \\\hline
     0.80  & baseline     & 0.152 {\scriptsize ($\pm$0.009)} & 0.351 {\scriptsize ($\pm$0.024)} & 0.201 {\scriptsize ($\pm$0.012)} & 0.295 {\scriptsize ($\pm$0.018)} & 1.128 {\scriptsize ($\pm$0.064)} \\
     0.80  & proposed &0.325 {\scriptsize ($\pm$0.012)} & 0.974 {\scriptsize ($\pm$0.019)} & 0.484 {\scriptsize ($\pm$0.015)} & 0.915 {\scriptsize ($\pm$0.053)} & 0.196 {\scriptsize ($\pm$0.130)} \\\hline
     0.70  & baseline      &0.158 {\scriptsize ($\pm$0.009)} & 0.323 {\scriptsize ($\pm$0.025)} & 0.199 {\scriptsize ($\pm$0.012)} & 0.277 {\scriptsize ($\pm$0.019)} & 1.176 {\scriptsize ($\pm$0.060)} \\
     0.70  & proposed & 0.346 {\scriptsize ($\pm$0.013)} & 0.902 {\scriptsize ($\pm$0.031)} & 0.493 {\scriptsize ($\pm$0.017)} & 0.875 {\scriptsize ($\pm$0.057)} & 0.254 {\scriptsize ($\pm$0.126)} \\\hline
     0.60  & baseline   &0.164 {\scriptsize ($\pm$0.010)} & 0.304 {\scriptsize ($\pm$0.027)} & 0.198 {\scriptsize ($\pm$0.013)} & 0.295 {\scriptsize ($\pm$0.022)} & 1.112 {\scriptsize ($\pm$0.069)} \\
     0.60  & proposed & 0.336 {\scriptsize ($\pm$0.011)} & 0.830 {\scriptsize ($\pm$0.035)} & 0.473 {\scriptsize ($\pm$0.016)} & 0.789 {\scriptsize ($\pm$0.068)} & 0.417 {\scriptsize ($\pm$0.146)} \\\hline
     0.50  & baseline     &0.173 {\scriptsize ($\pm$0.009)} & 0.298 {\scriptsize ($\pm$0.026)} & 0.200 {\scriptsize ($\pm$0.012)} & 0.274{\scriptsize ($\pm$0.023)} & 1.151 {\scriptsize ($\pm$0.070)}    \\
     0.50  & proposed & 0.320 {\scriptsize ($\pm$0.015)} & 0.691 {\scriptsize ($\pm$0.040)} & 0.429 {\scriptsize ($\pm$0.019)} & 0.699 {\scriptsize ($\pm$0.062)} & 0.564 {\scriptsize ($\pm$0.143)} \\\hline
    \end{tabular} }
  \end{center}
\end{table}

\begin{figure}[tb]
  \centering
  \begin{tabular}{c@{\ \ \ \ \ }c@{\ \ \ \ \ }c}
  \includegraphics[width=3cm]{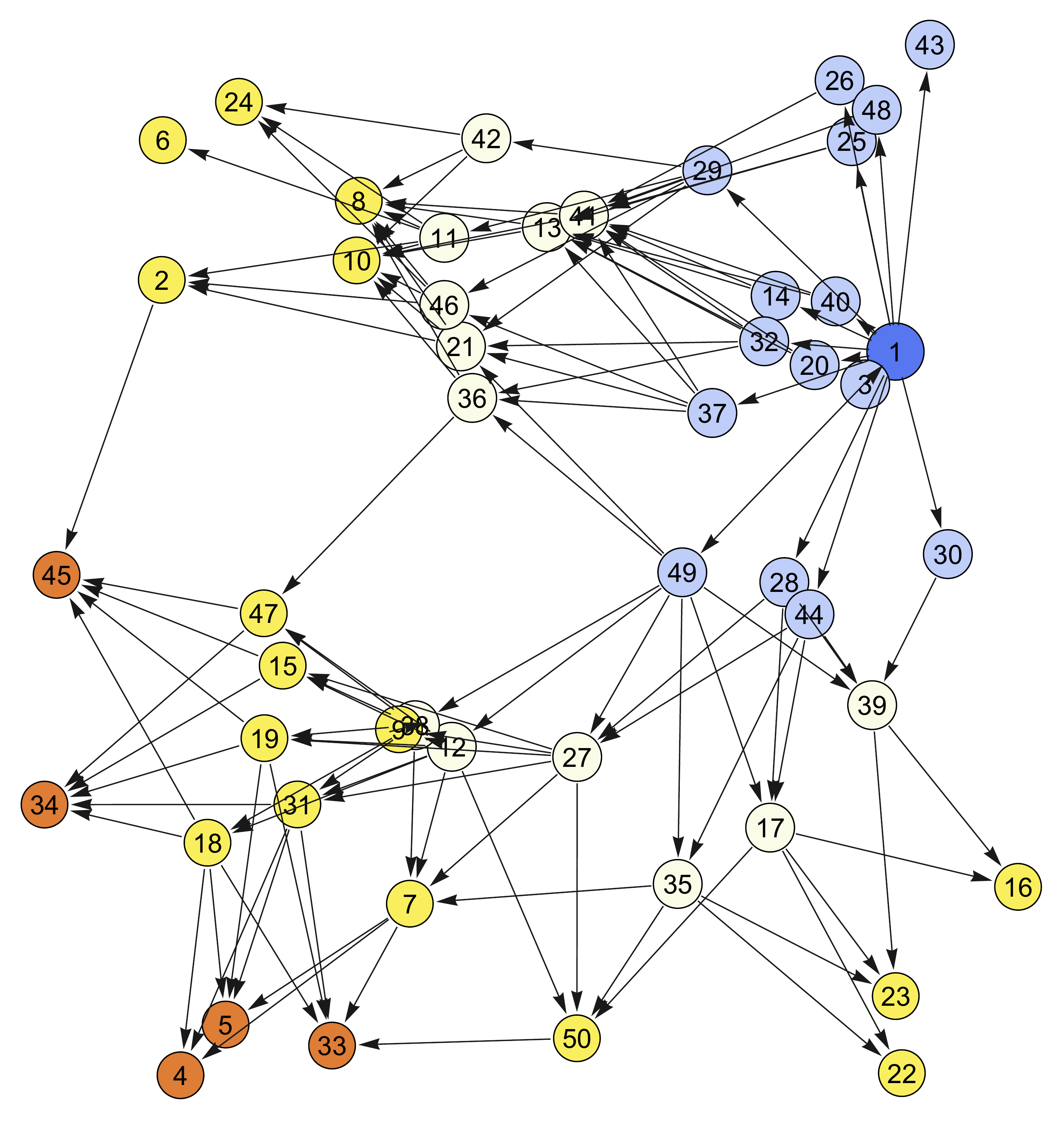}& \includegraphics[width=4cm]{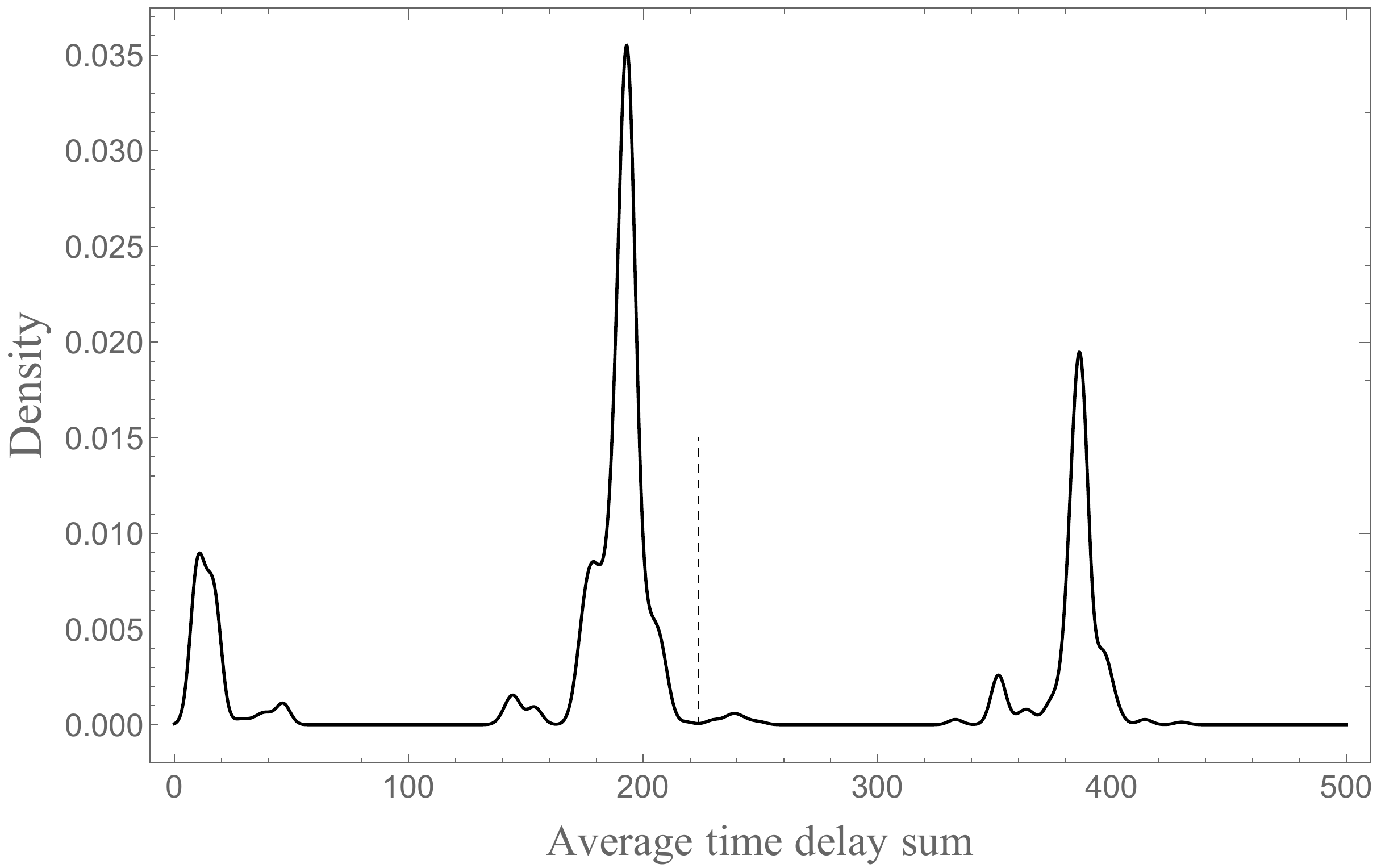} &\includegraphics[width=3cm]{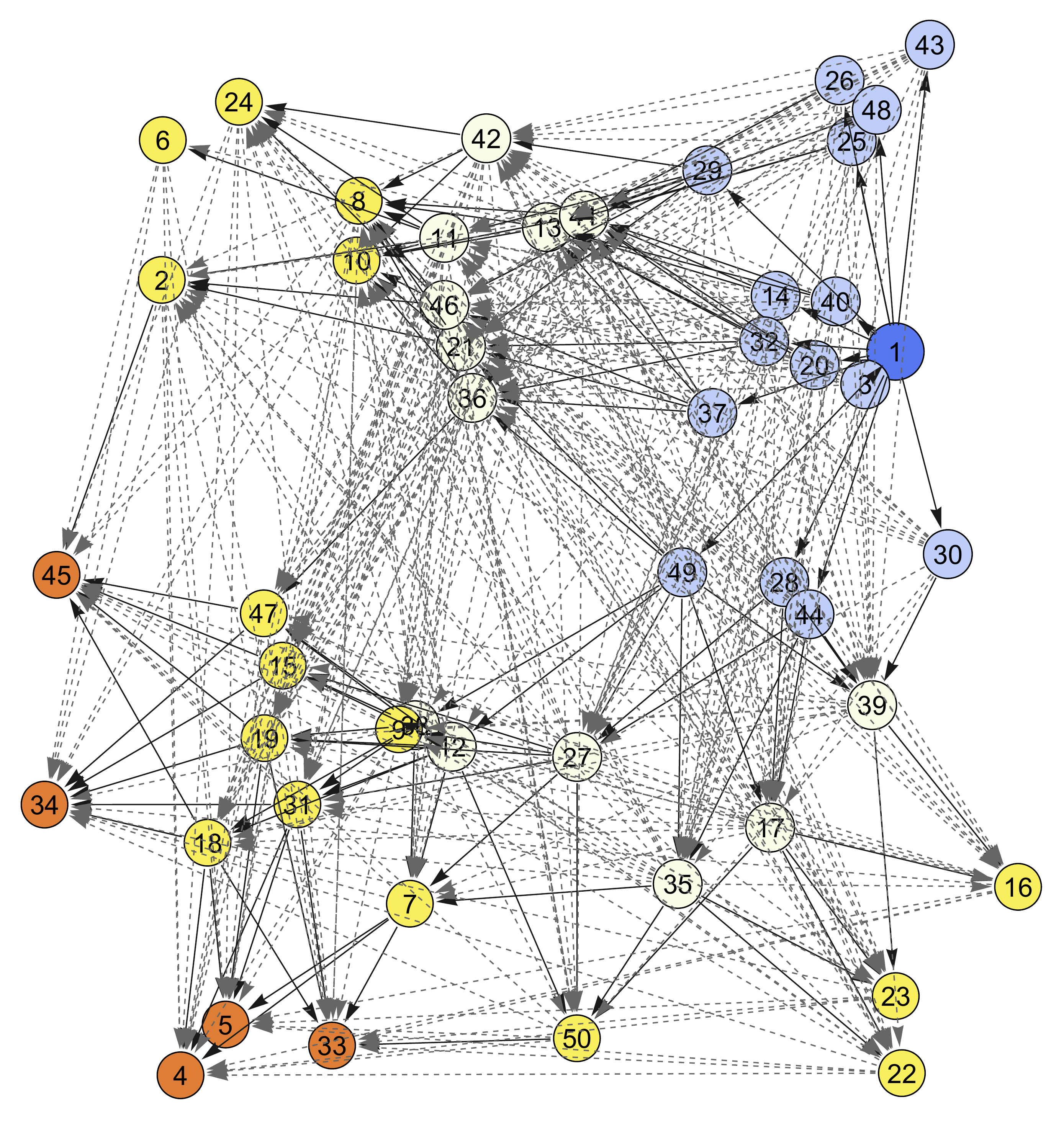}\\
  $G_{\mathbb{B}}(V,E)$ & &$G_{\mathbb{B}}(V,\hat{E})$\\
  \end{tabular}
  \caption{[Left] The ground truth graph $G_{\mathbb{B}}(V,E)$. [Center] The probability density of average time delay sum estimated by kernel density estimation for one of the synthetic datasets. [Right] The estimated graph $G_{\mathbb{B}}(V,\hat{E})$ by the proposed method for one of the ten datasets with $p=0.95$. Each individual $i$ is located at $r_i$.
    In $G_{\mathbb{B}}(V,\hat{E})$, solid arrows are edges in $E$ and dashed arrows are edges in $\hat{E}\setminus E$.
    Note that $\hat{E}$ includes $E$ (recall $1.0$). Each individual's color indicates its belonging layer: dark blue, blue, light blue, yellow, orange and red 
    individuals belong to $V^{\hat{E}}_0$, $V^{\hat{E}}_1$, $V^{\hat{E}}_2$, $V^{\hat{E}}_3$, $V^{\hat{E}}_4$, and $V^{\hat{E}}_5$, respectively. 
}\label{fig:ex1}
\end{figure}
Performance comparison with the baseline method by our evaluation measures is shown Table~\ref{tab:result}. 
The proposed method also outperformed the baseline method in all the measures.
Precisions of both the methods are low compared to their recalls, that is due to correct edge (directly affecting edge) definition: location information is used to define the ground truth graph edges but such information cannot be used in this experimental setting.
Our method successfully estimates each individual's belonging layer with higher LA and lower MLD when $p$ is around $1$
and keeps LA about 0.7 even for $p=0.5$.

The estimated graph $G(V,\hat{E})$ by the proposed method for one of the datasets with $p=0.95$ is shown on the right of Fig.~\ref{fig:ex1}.
For the dataset, parameter $\theta$ is set to $223.48$ from the estimated distribution (the center of Fig.~\ref{fig:ex1}).
There are many falsely detected edges but all the edges in $E$ are correctly detected keeping the layer structure.

\subsection{Application to Real Datasets}
\subsubsection{Stock Price Analysis}

\begin{figure}[t]
\noindent
\includegraphics[width=4cm]{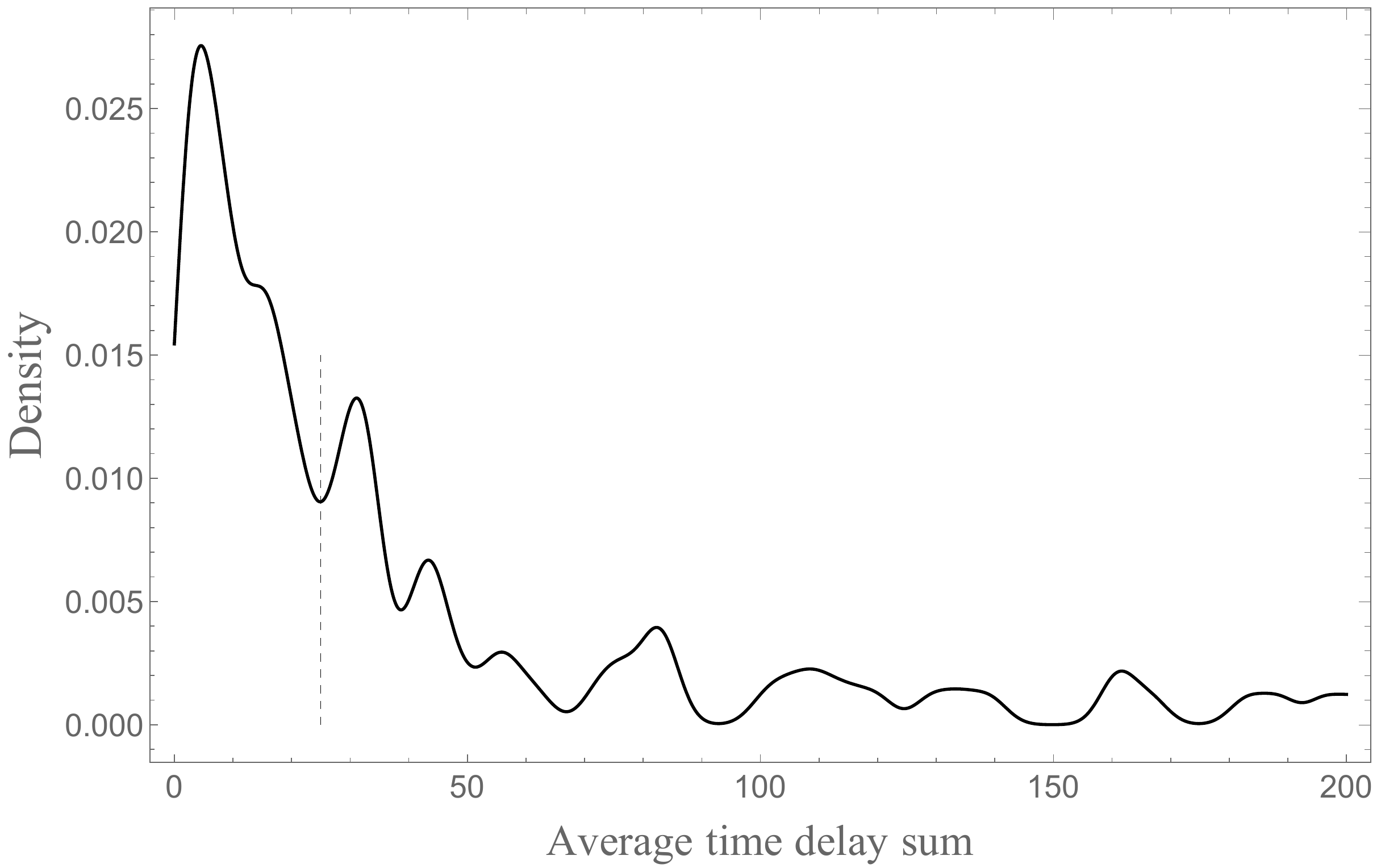}  
\includegraphics[width=4.5cm]{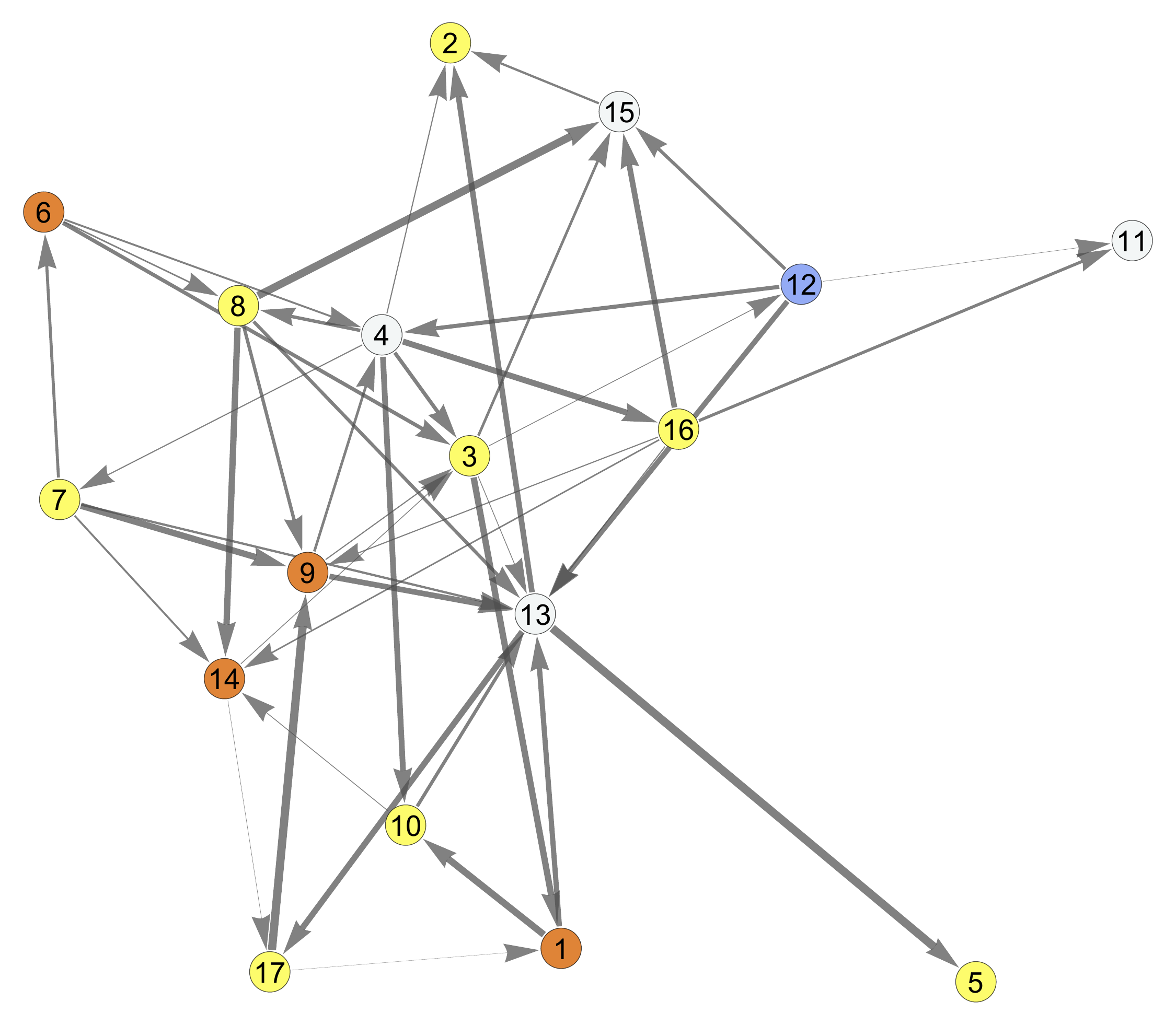}
\begin{minipage}[b]{3cm}
{\tiny
\begin{tabular}{l}
1 foods\\
2 energy resources\\ 
3 construction \& materials\\ 
4 raw materials \& chemicals\\ 
5 pharmaceutical\\
6 automobiles \&\\
\phantom{6} transportation equipment\\ 
7 steel \& nonferrous metals\\
8 machinery\\
9 electric appliances \&\\
\phantom{9} precision instruments\\ 
10 IT \& services, others\\
11 electric power \& gas\\
12 transportation \& logistics\\
13 commercial \& wholesale trade\\
14 retail trade\\
15 banks\\
16 financials (ex banks)\\
17 real estate\\
\end{tabular}
}
\end{minipage}
\caption{[Left] Probability density of average time delay sum estimated by kernel density estimation for stock market datasets. We used Gaussian kernel with a bandwidth of three. The dashed line ($\theta = 24.91$) indicates a threshold value adopted by our method. [Right] The estimated stock price propagation graph  by the proposed method. The number in each node shows the sector. Each node's color indicates its belonging layer: dark blue,  light blue, yellow and orange
    nodes belong to $V^{\hat{E}}_0$, $V^{\hat{E}}_1$, $V^{\hat{E}}_2$, and $V^{\hat{E}}_3$, respectively. The thickness of an edge shows the size of the average time delay; the thicker the edge, the longer the delay.
}\label{fig:stock_graph}
\end{figure}
\begin{figure}[t]
\noindent
\begin{tabular}{@{\ \ \ }p{4cm}@{\ \ \ }p{7cm}}
  \begin{minipage}[b]{4cm}
    \includegraphics[width=4cm]{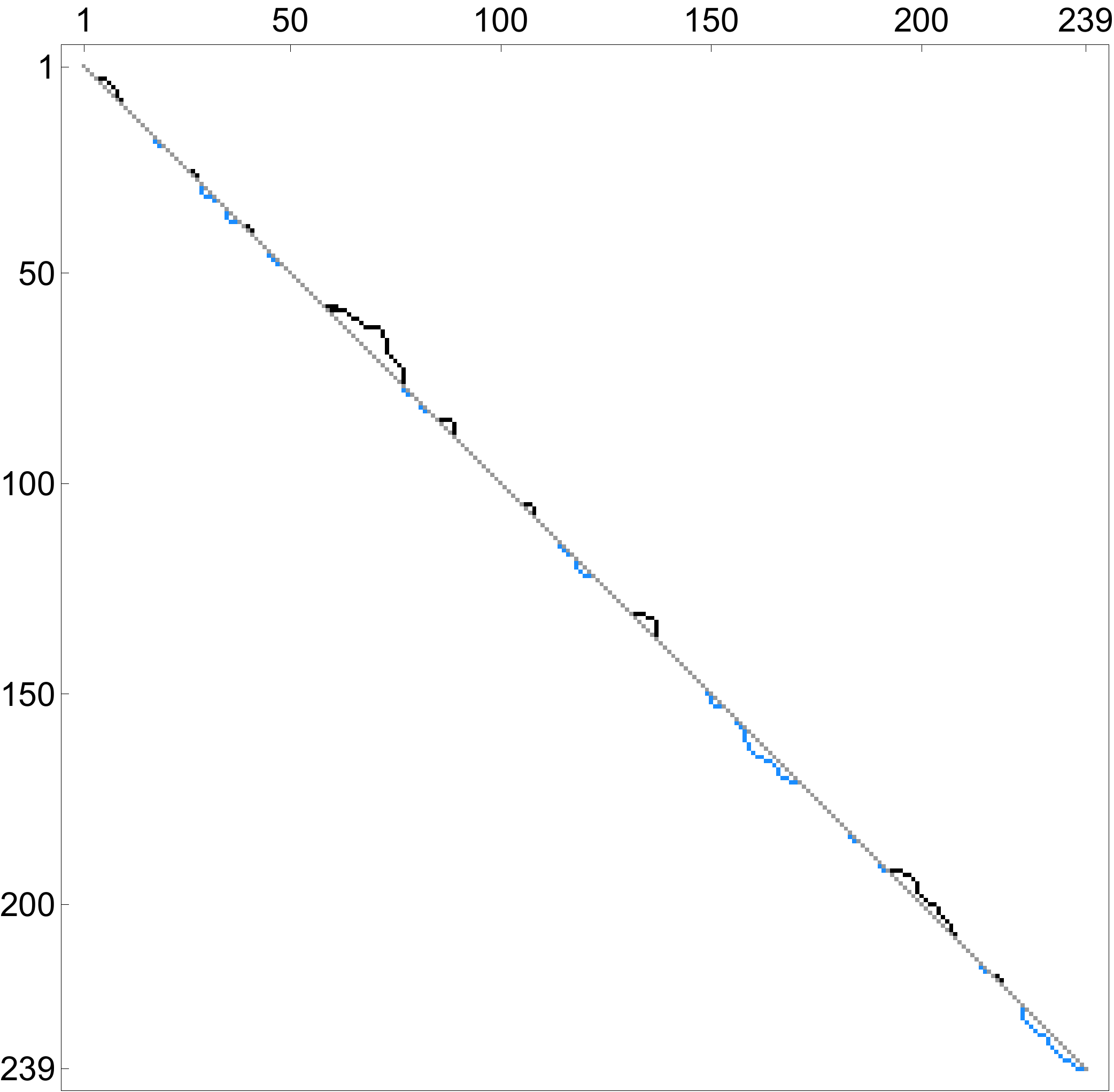}\\
    \end{minipage}
    &
  \includegraphics[width=7cm]{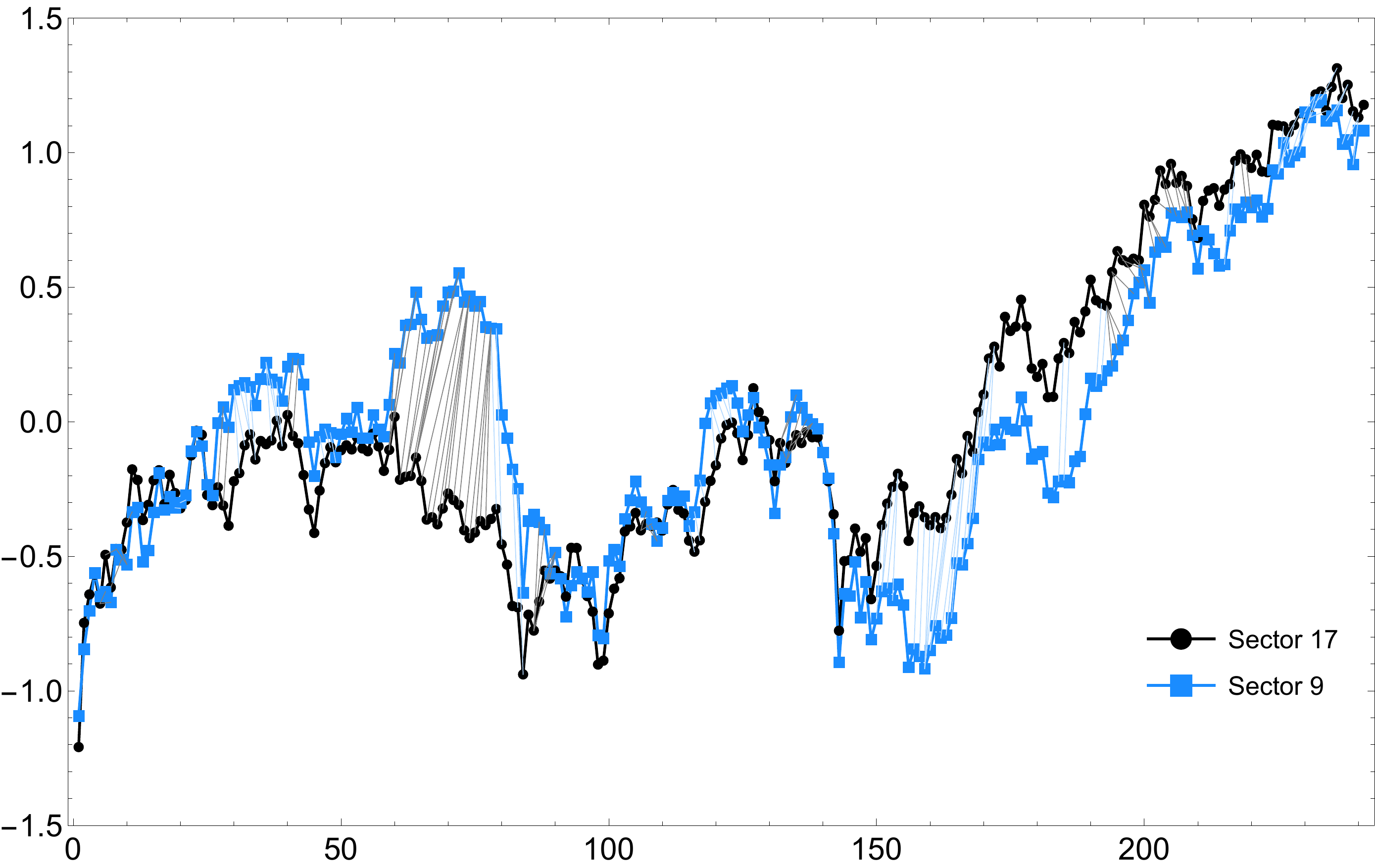}
\end{tabular}
\begin{minipage}{\textwidth}
{\scriptsize
In the right figure, the horizontal axis is time, and the vertical axis is standardized stock price.
Lines between $s'_{9}$ and $s'_{17}$ indicate correspondences between the estimated stock price derivative time series $s_{9}$ and $s_{17}$ in the minimum cost alignment,
and are drawn between points $(\pi_{9}^{-1}(k),s'_{9}[\pi_{9}^{-1}(k)])$ and $(\pi_{17}^{-1}(k),s'_{17}[\pi_{17}^{-1}(k)])$ for shifted aligned positions $k$. The gray (light blue) lines indicate that the sector 9 (sector 17) follows the sector 17 (sector 9).}
\end{minipage}
\caption{[Left] The minimum cost alignment path between $s_{9}$ and $s_{17}$. [Right] Graph of $s'_{9}$ and $s'_{17}$ with their matched positions. The average time delay sum of $s_9$ from $s_{17}$ is $22.0$. 
}\label{fig:sector_ex}
\end{figure}

We report our analysis of stock price propagation by the proposed method.
We used the datasets of stock price time series of 2145 companies listed on the first section of the Tokyo Stock Exchange for the period from 4th January to 30th December in 2019. The set of the listed companies is partitioned into 17 sectors by TOPIX-17 series\footnote{https://www.jpx.co.jp/english/markets/indices/line-up/files/e\_fac\_13\_sector.PD}. 
The given time series $p_j[t]$ ($t=0,\dots,240$) is the sequence of the opening stock price of company $j$ on $t$th day for $j=1,\dots,2145$. 
We standardized each time series $p_j$ to $p'_j$ so that $p'_j[t]$ ($t=0,\dots,240$) have mean zero and standard deviation one.
The time series $s'_i[t]$ ($t=0,\dots,240$) is the standardized sequence of the opening stock price on $t$th day averaged over companies in sector $i$ for $i=1,\dots,17$.
Then, $s_i[t]$ ($t=1,\dots,239$), which is an estimated derivative of $s'_i$ at time $t$, is calculated by equation $s_i[t] = \frac{(s'_i[t]-s'_i[t-1]) + (s'_i[t+1]-s'_i[t-1])/2}{2}$.
The right figure of Fig.~\ref{fig:stock_graph} shows the estimated propagation graph among 17 sectors by the proposed method for threshold $\theta = 24.91$, which is determined from estimated distribution of average time delay sum (the left graph of Fig.~\ref{fig:stock_graph}). As an example, Fig.~\ref{fig:sector_ex} shows the minimum cost path between the time series $s_{9}$ and $s_{17}$, and the graph of $s'_{9}$ and $s'_{17}$ with their matched positions.
You can see that $s_9$ (derivative of $s'_9$) follows $s_{17}$ during two long time periods $[59,77]$ and $[193,208]$ with small time delays. 

\begin{figure}[tb]
 \centering
  \includegraphics[width=5cm]{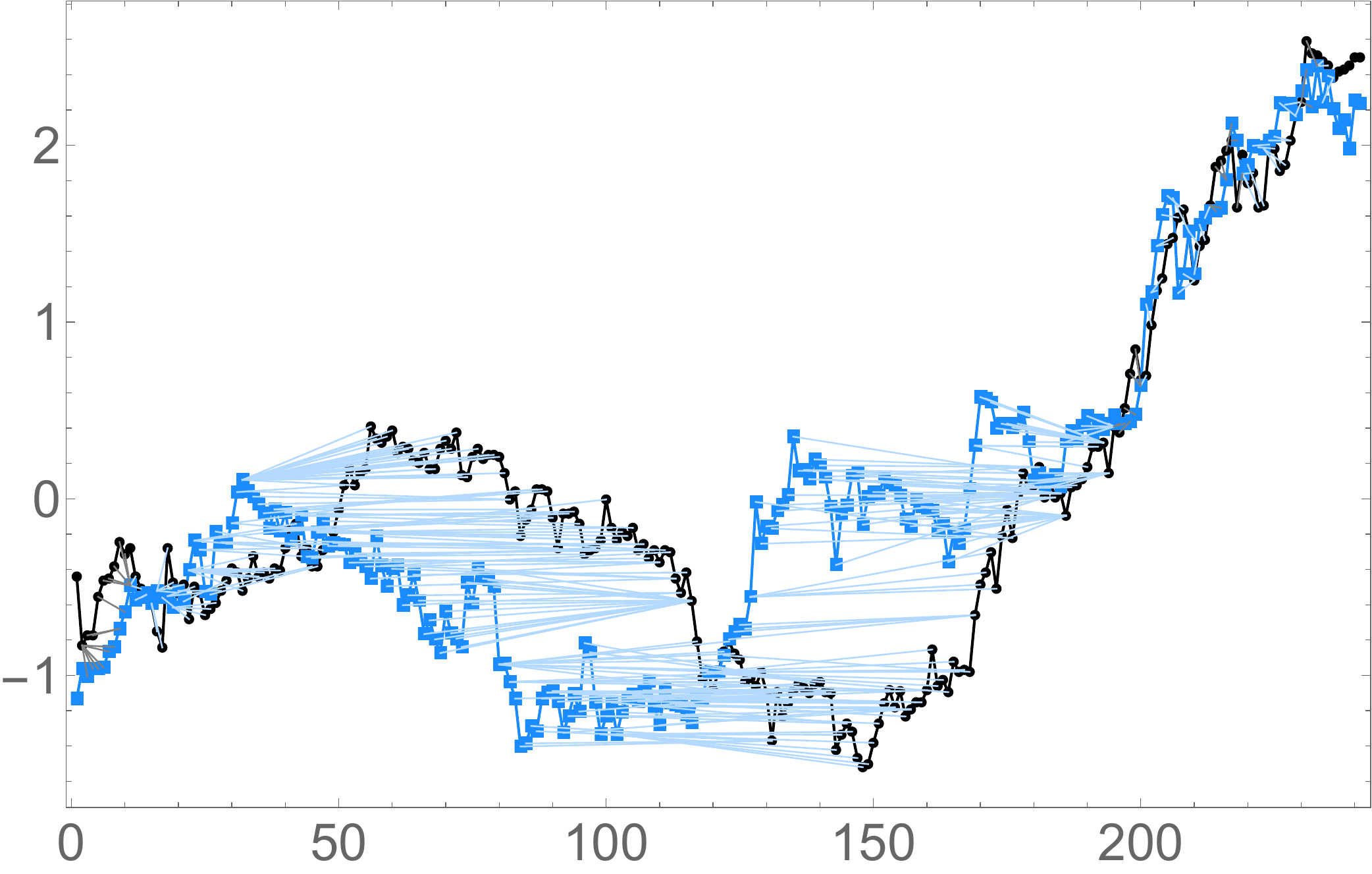}\\
  \caption{The standardized sequences of the opening price for NAGAWA (black) and KYOKUTO BOEKI KAISHA (blue). Lines between them indicate their corresponding positions. The horizontal axis is time, and the vertical axis is standardized stock price.
  NAGAWA looks following to KYOKUTO BOEKI KAISHA with large time delay during time period between 60 and 190.}\label{delay_example}
\end{figure}

Among the set of pairs of individual stocks, stock pairs that have clearer leader-follower relationship can be found.
Fig.~\ref{delay_example} shows the standardized sequences of the opening price for one of those pairs (``NAGAWA'', ``KYOKUTO BOEKI KAISHA'').
with the lines connecting corresponding points between them.
In the figure, you can see that black stock (NAGAWA) follows blue stock (KYOKUTO BOEKI KAISHA) with large time delay during period between 60 and 190.

\subsubsection{Cell's Firing Analysis} \label{cellfiring}

We applied our method to firing state propagation of biological cells.
The dataset is composed of $250$-frame $\{0,1\}$-state and 2D-location sequences of $172$ cells, where states $1$ and $0$ represent firing and not firing, respectively.
Our method uses state sequences alone and location sequence is used only for result visualization.

%

We used the data of 144 cells except for 28 cells which could not be measured properly due to noise. 
From the set of 144 binary sequences with length 250, we extracted 4 datasets $I_1, I_2, I_3$ and $I_4$, which is composed of 144 length-$100$ consecutive subsequences starting at frame $t=1, 51,101$ and $151$, respectively, of the original length-$250$ sequences.

\begin{figure}[h]
 \centering
  \begin{tabular}{cc}
  \includegraphics[width=4.5cm]{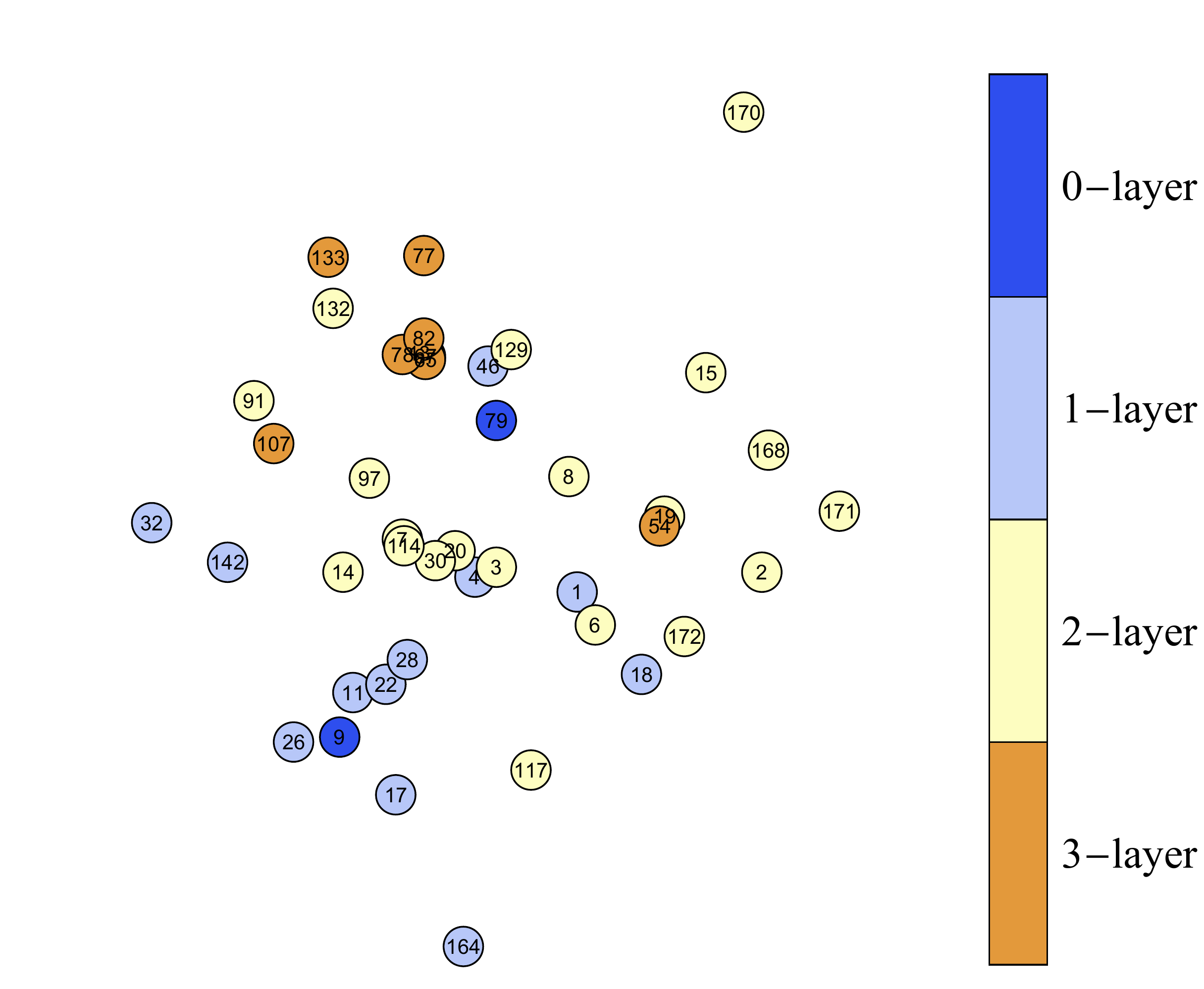}& \includegraphics[width=4.5cm]{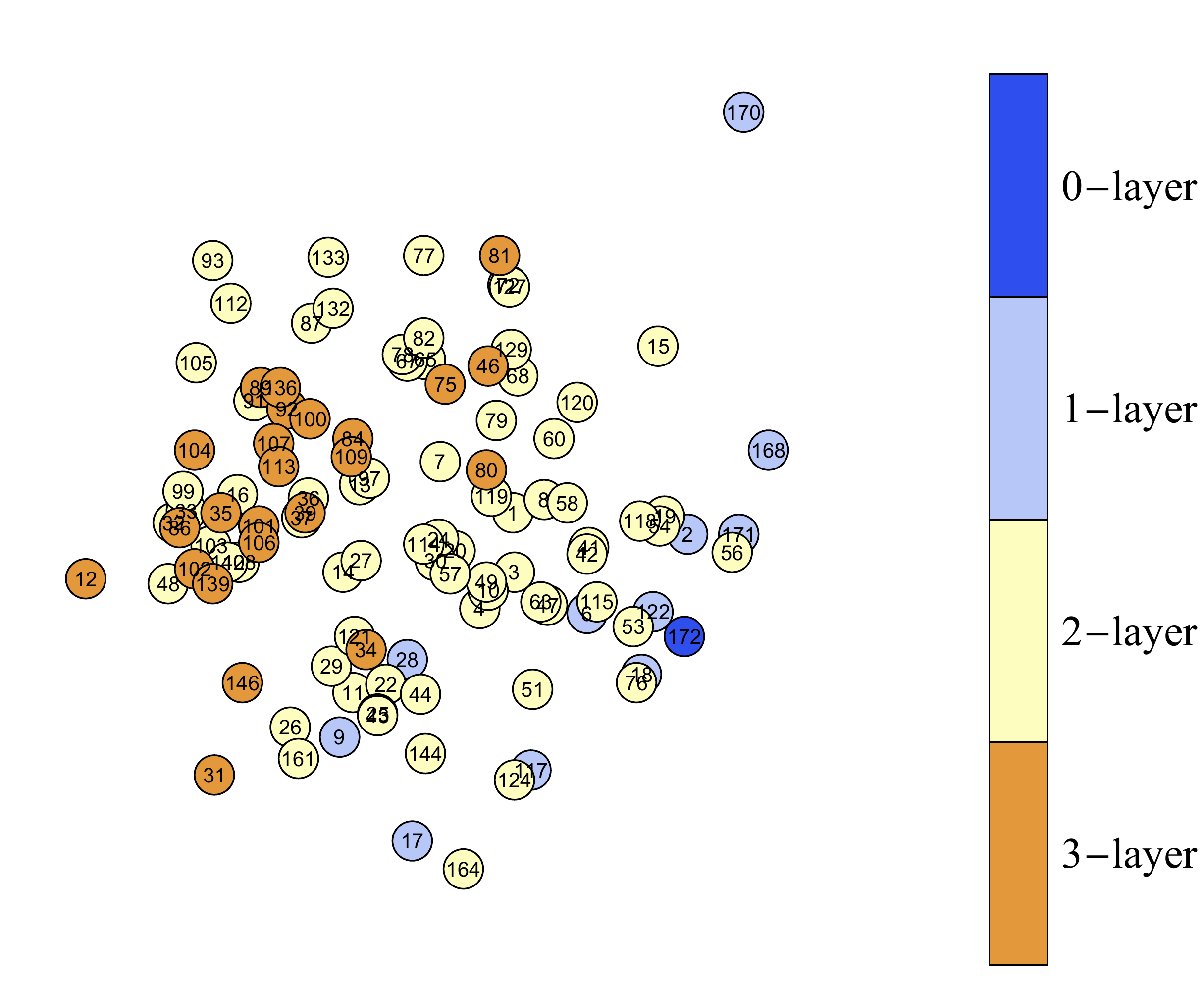}\\
  \includegraphics[width=4.5cm]{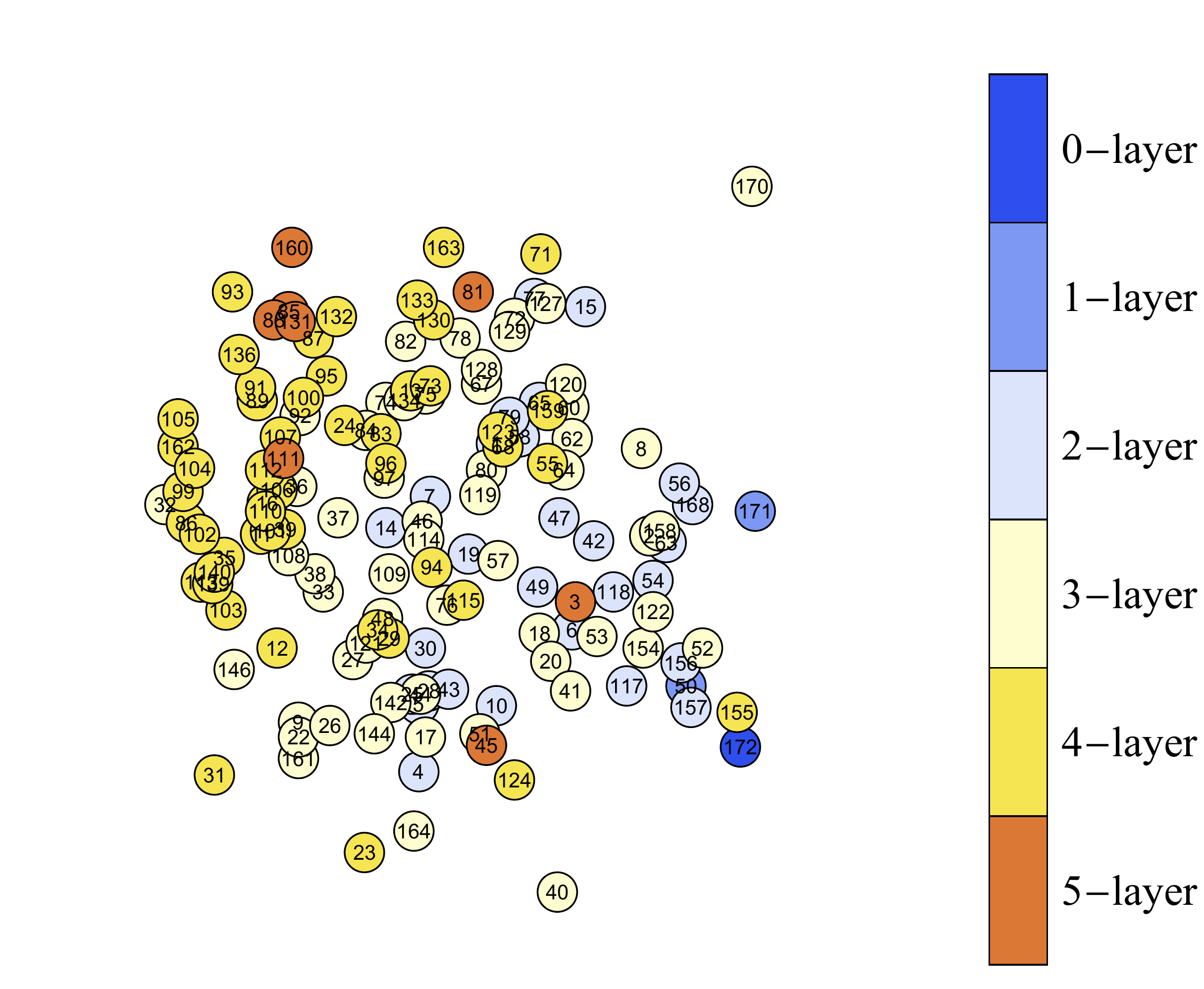}& \includegraphics[width=4.5cm]{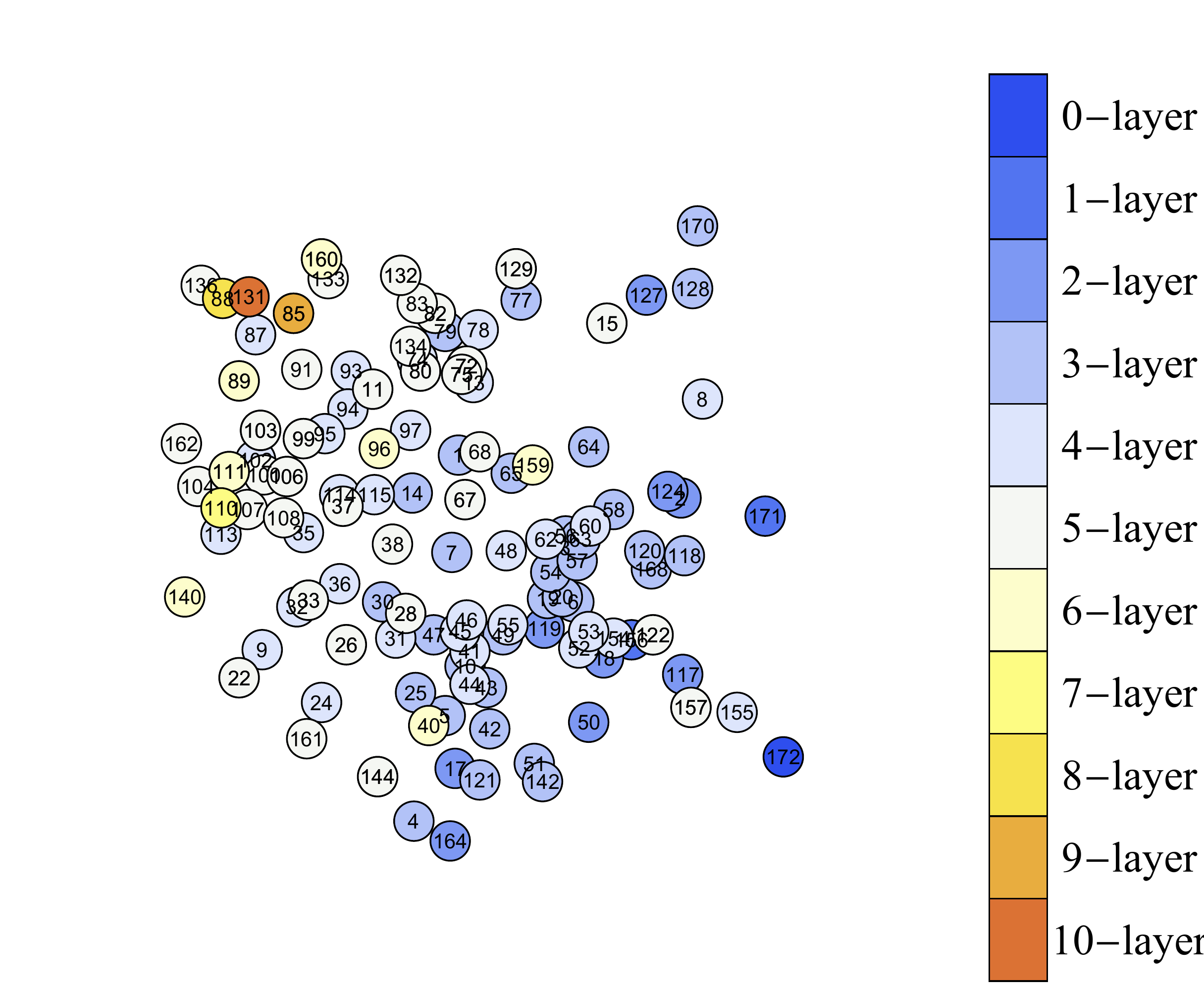}\\
  \end{tabular}
\caption{Layer partitions of the estimated propagation graphs for $I_1$(top left, cell location: $t=50$), $I_2$(top right, cell location: $t=100$), $I_3$(bottom left, cell location: $t=150$), $I_4$(bottom right, cell location: $t=200$).
We set $\theta = 65.53(I_1), 54.52(I_2), 12.29(I_3), 18.71(I_4)$.}\label{LP}
\end{figure}

\begin{figure}[h]
  \centering
  \begin{tabular}{cc}
  \includegraphics[width=4.5cm]{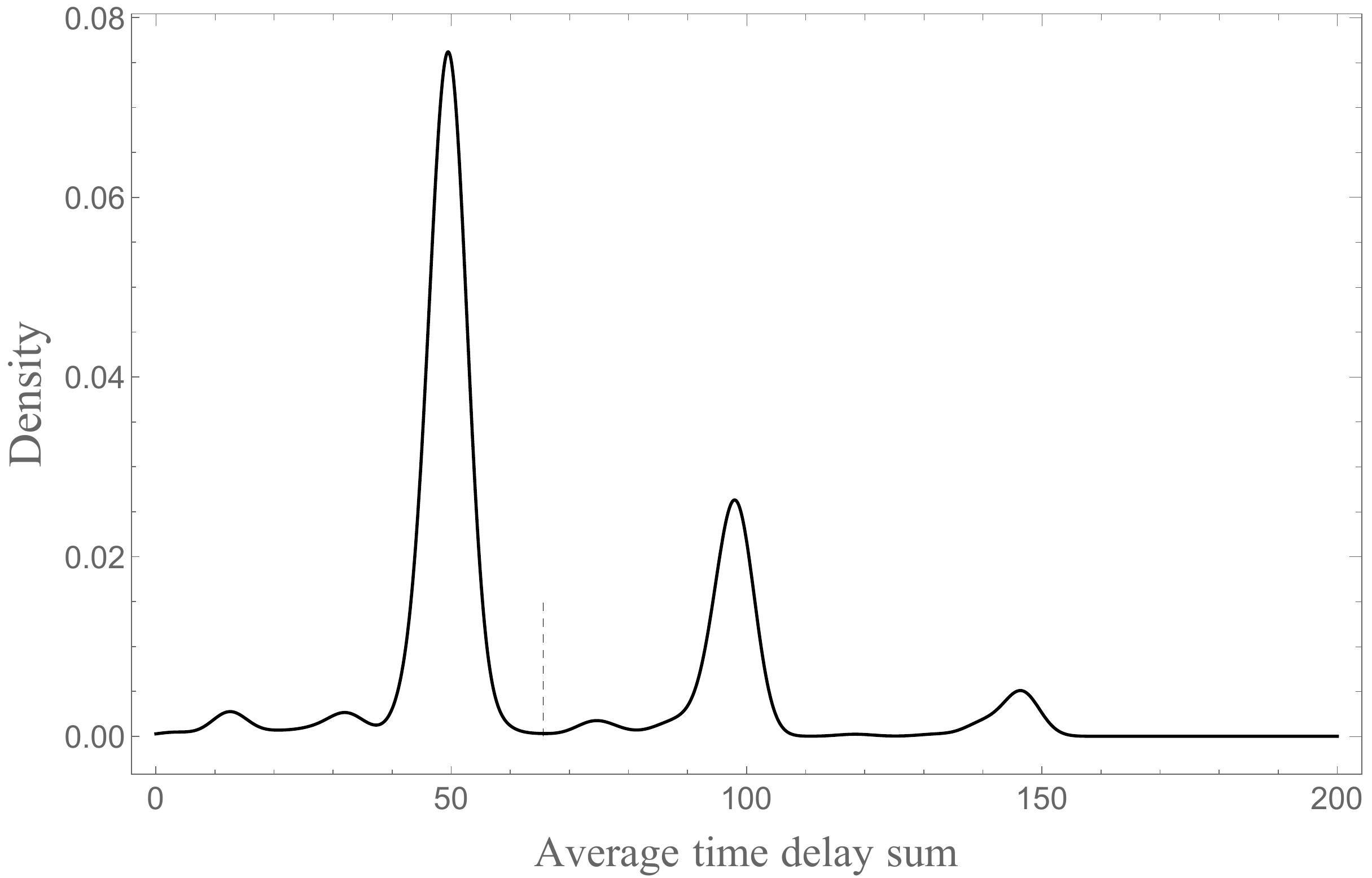}& \includegraphics[width=4.5cm]{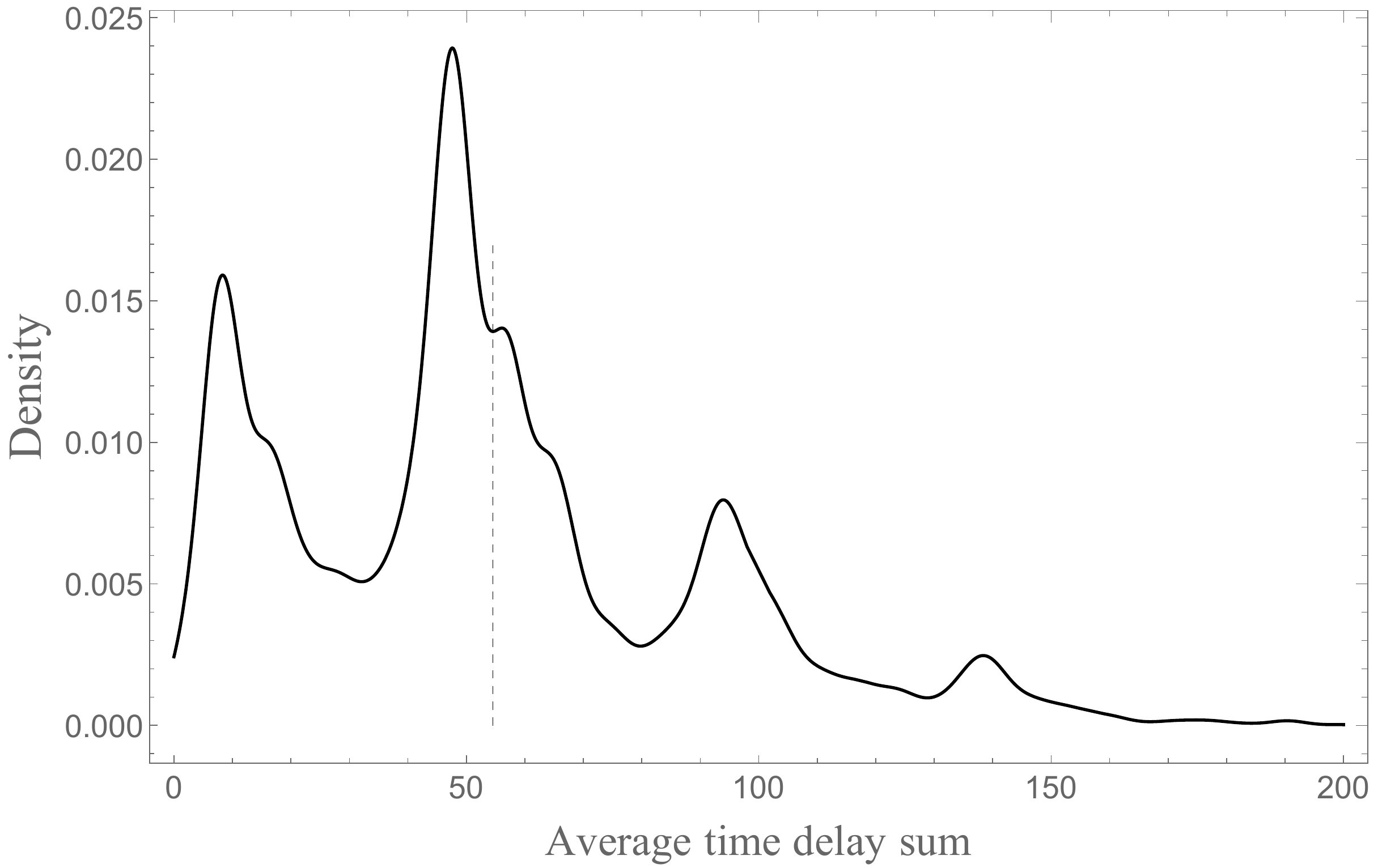}\\
  \includegraphics[width=4.5cm]{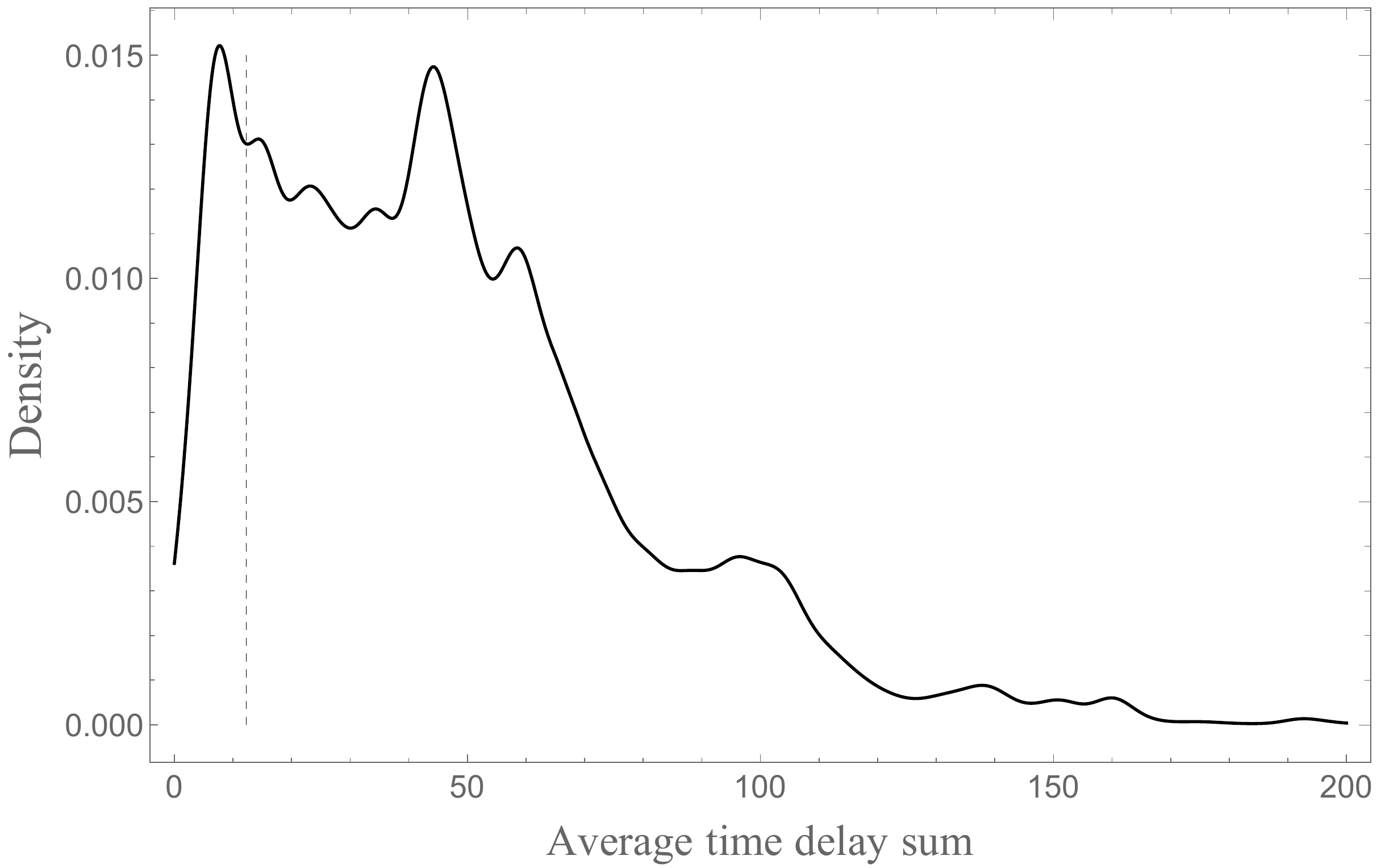}& \includegraphics[width=4.5cm]{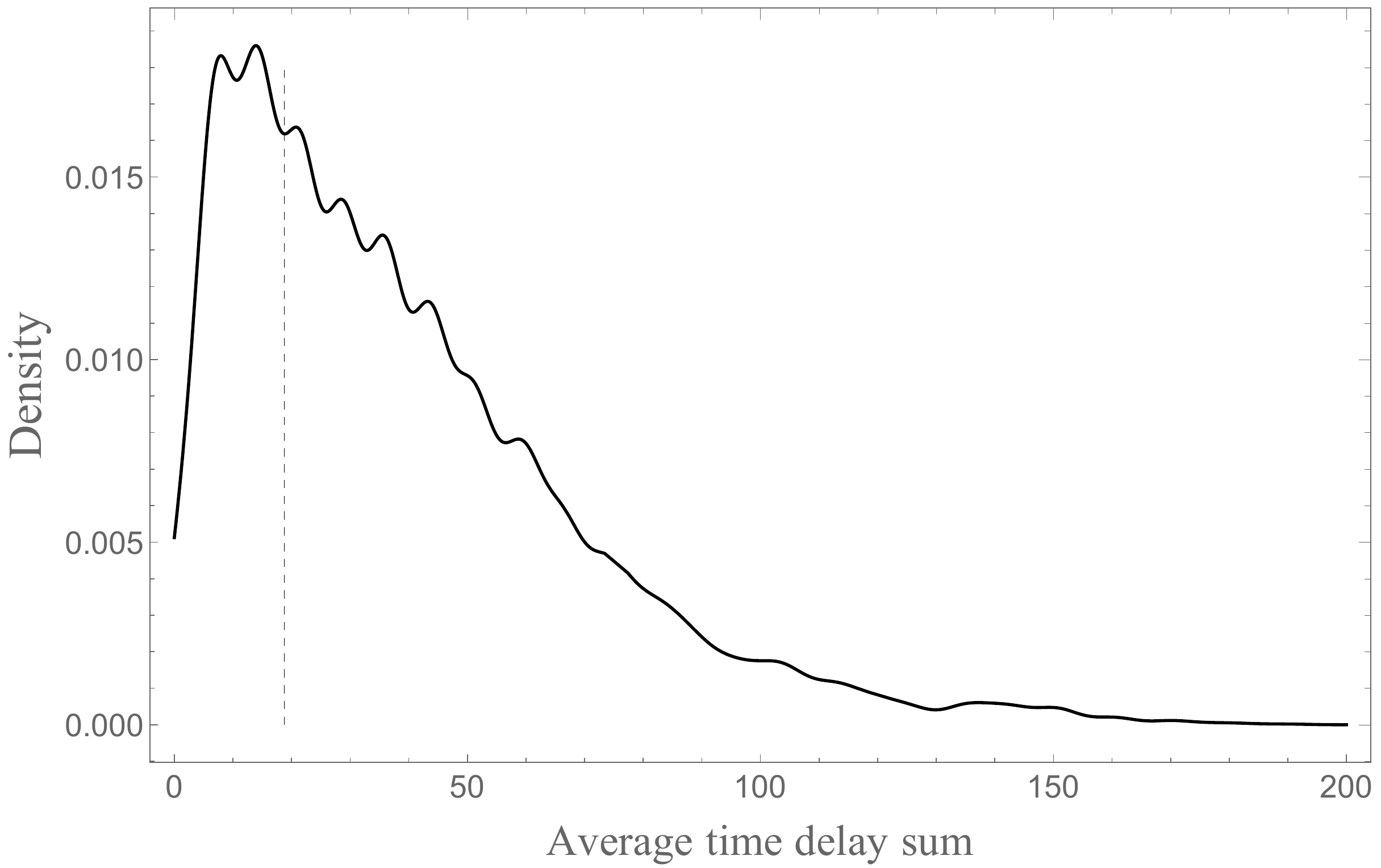}\\
  \end{tabular}
  \caption{Probability density of average time delay sum estimated by kernel density estimation for each dataset; top left is $I_1$, top right is $I_2$, bottom left is $I_3$, and bottom right is $I_4$. We used Gaussian kernel with a bandwidth of three. The dashed line in each graph is the adopted threshold value.}\label{fig:kernel_firing}
\end{figure}

\begin{figure}[h]
 \centering
  \begin{tabular}{ccc}
  \includegraphics[width=3.5cm]{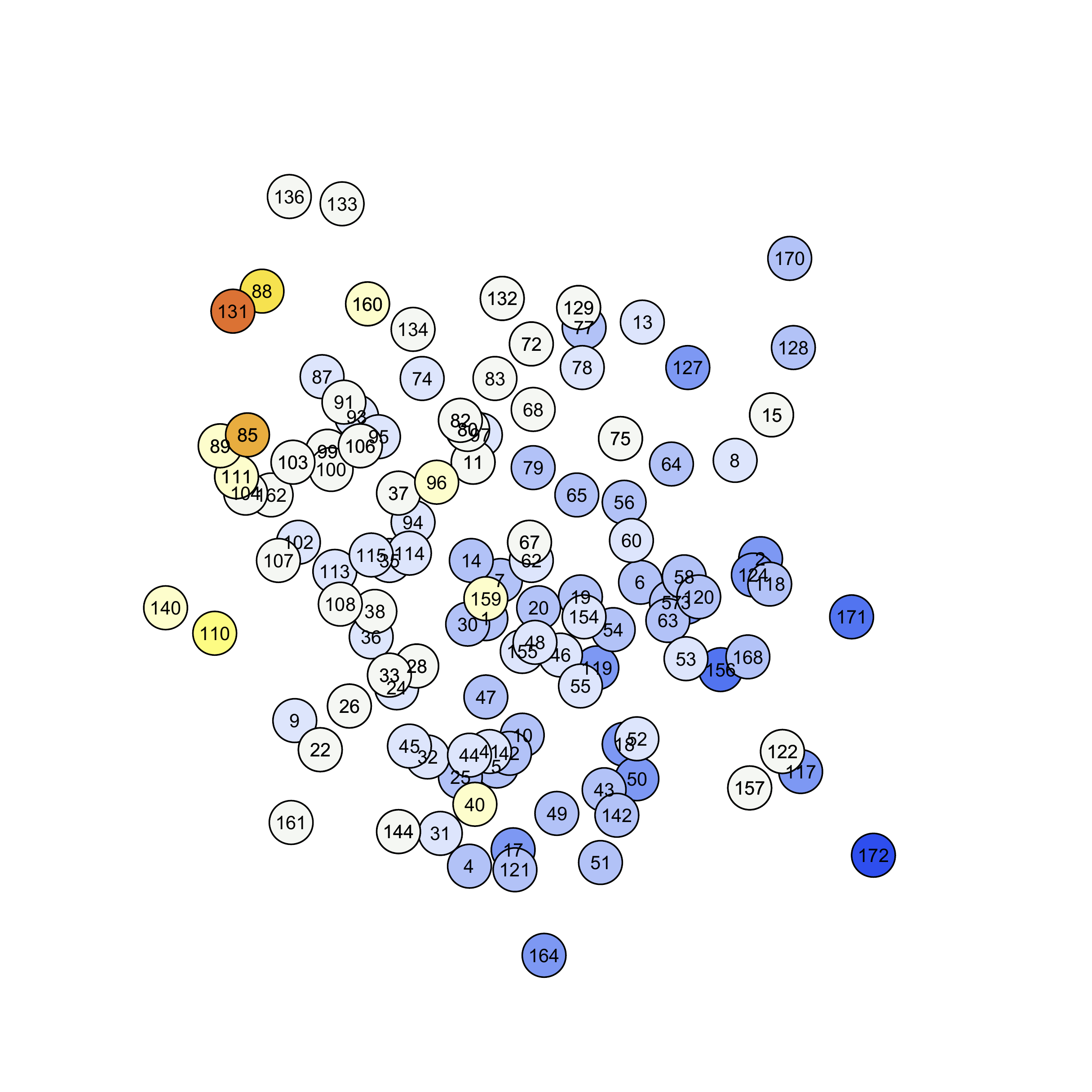}& \includegraphics[width=3.5cm]{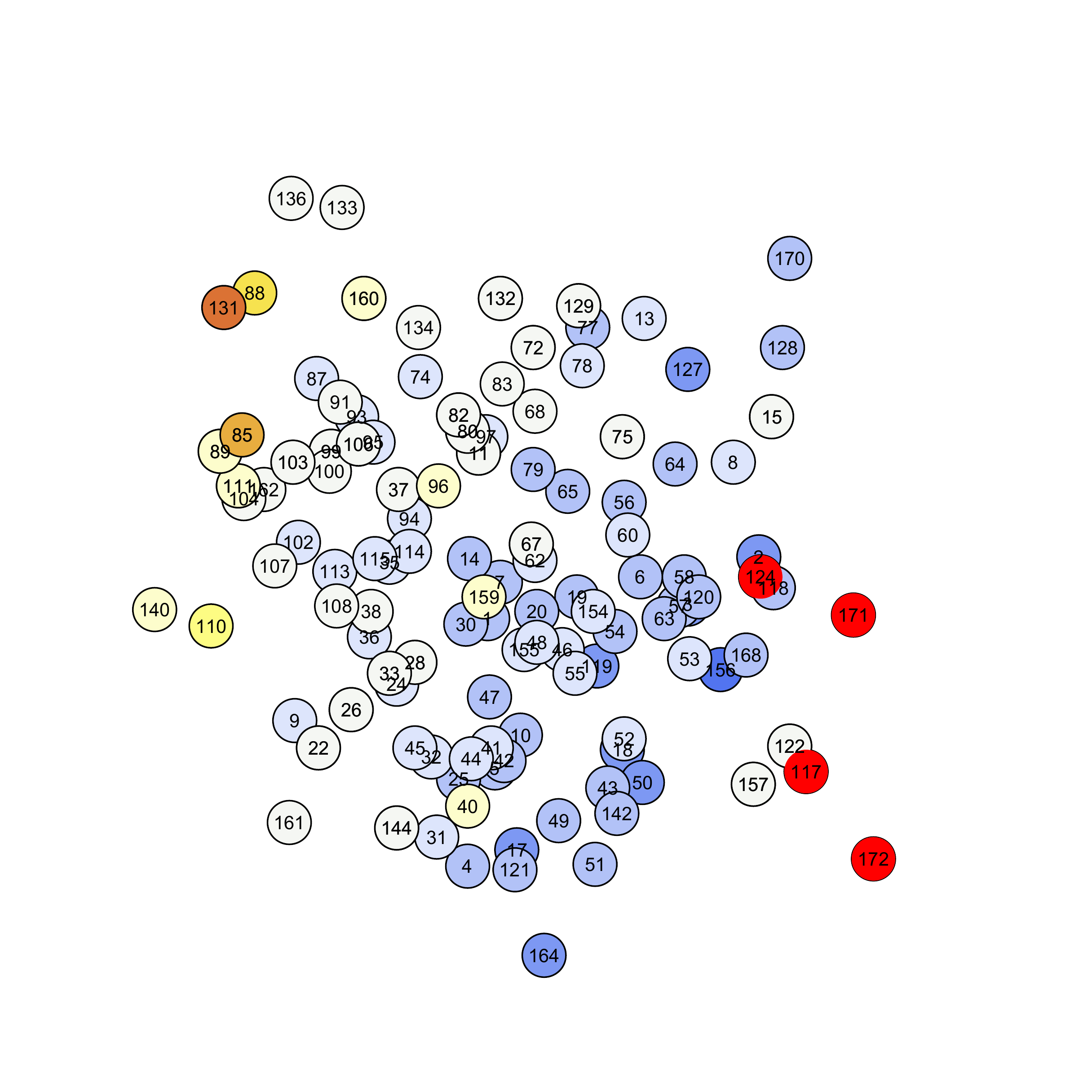}& \includegraphics[width=3.5cm]{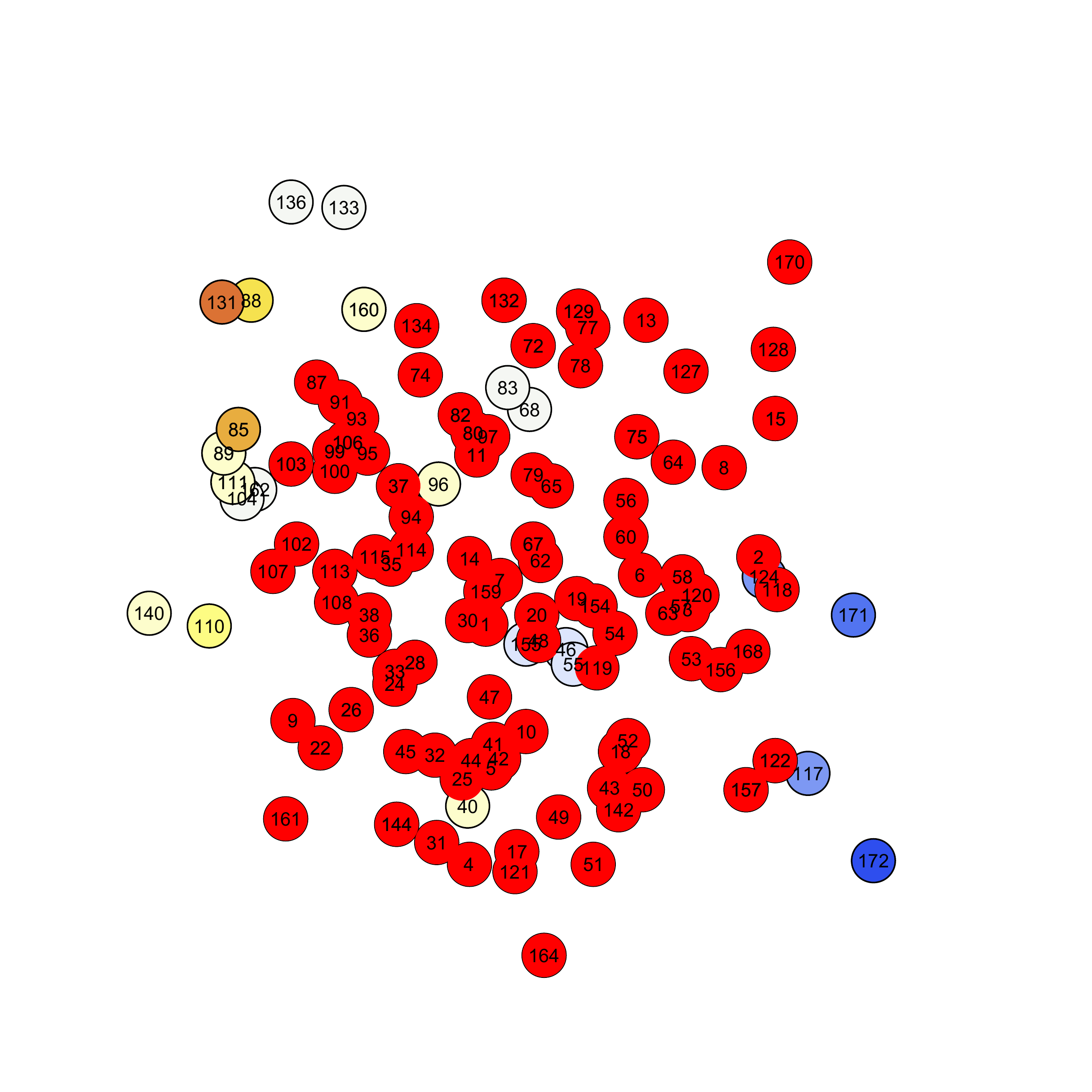}\\
  \includegraphics[width=3.5cm]{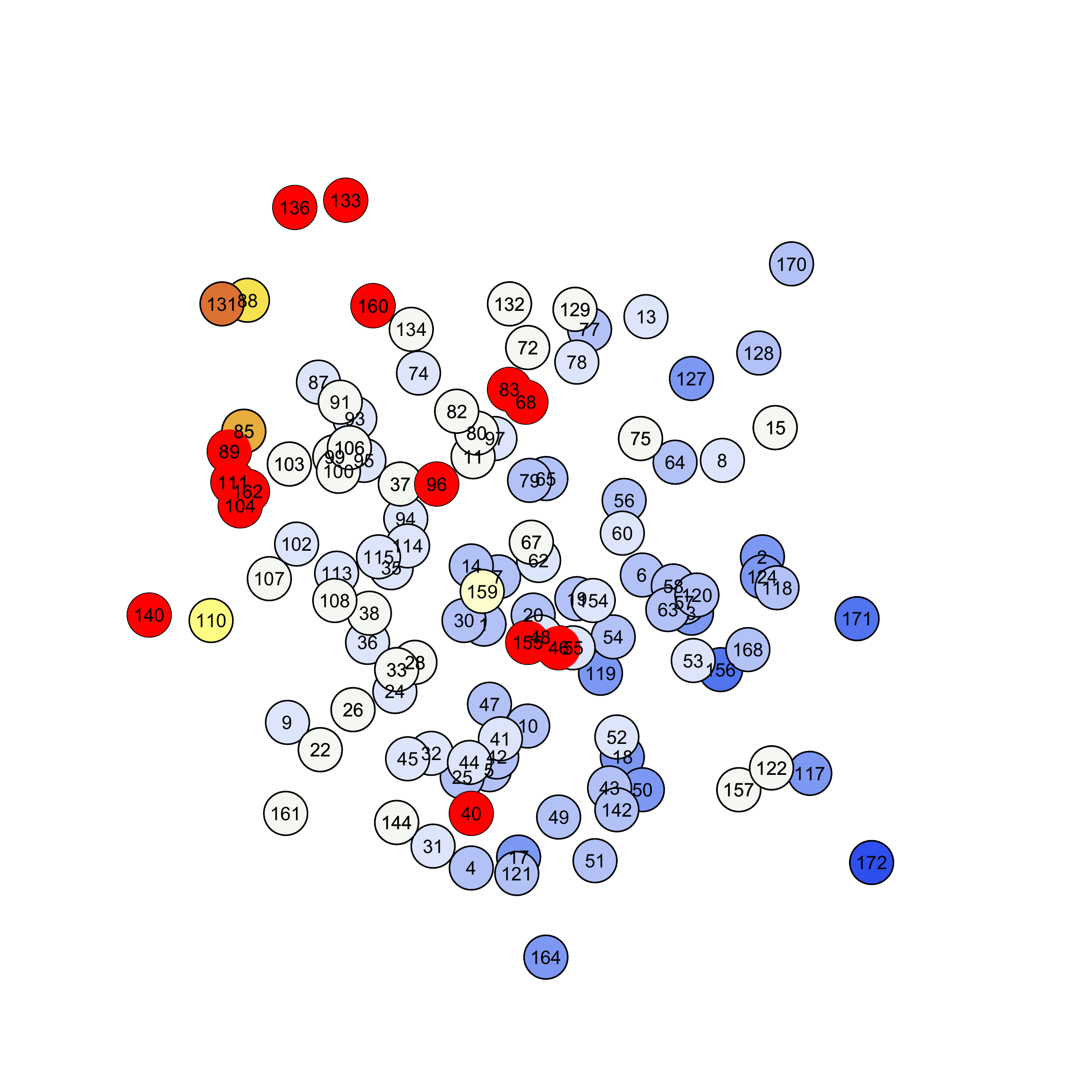}& \includegraphics[width=3.5cm]{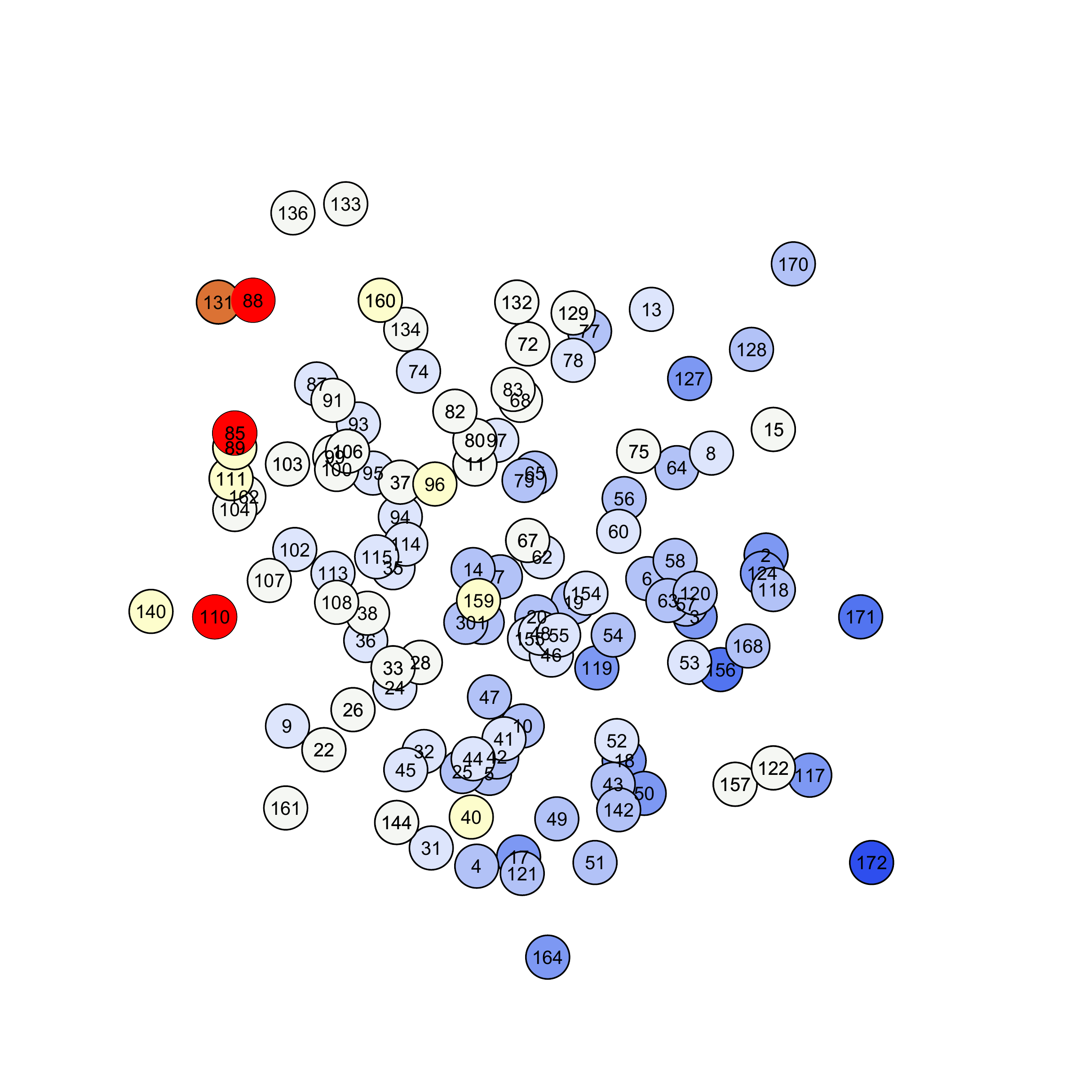}& \includegraphics[width=3.5cm]{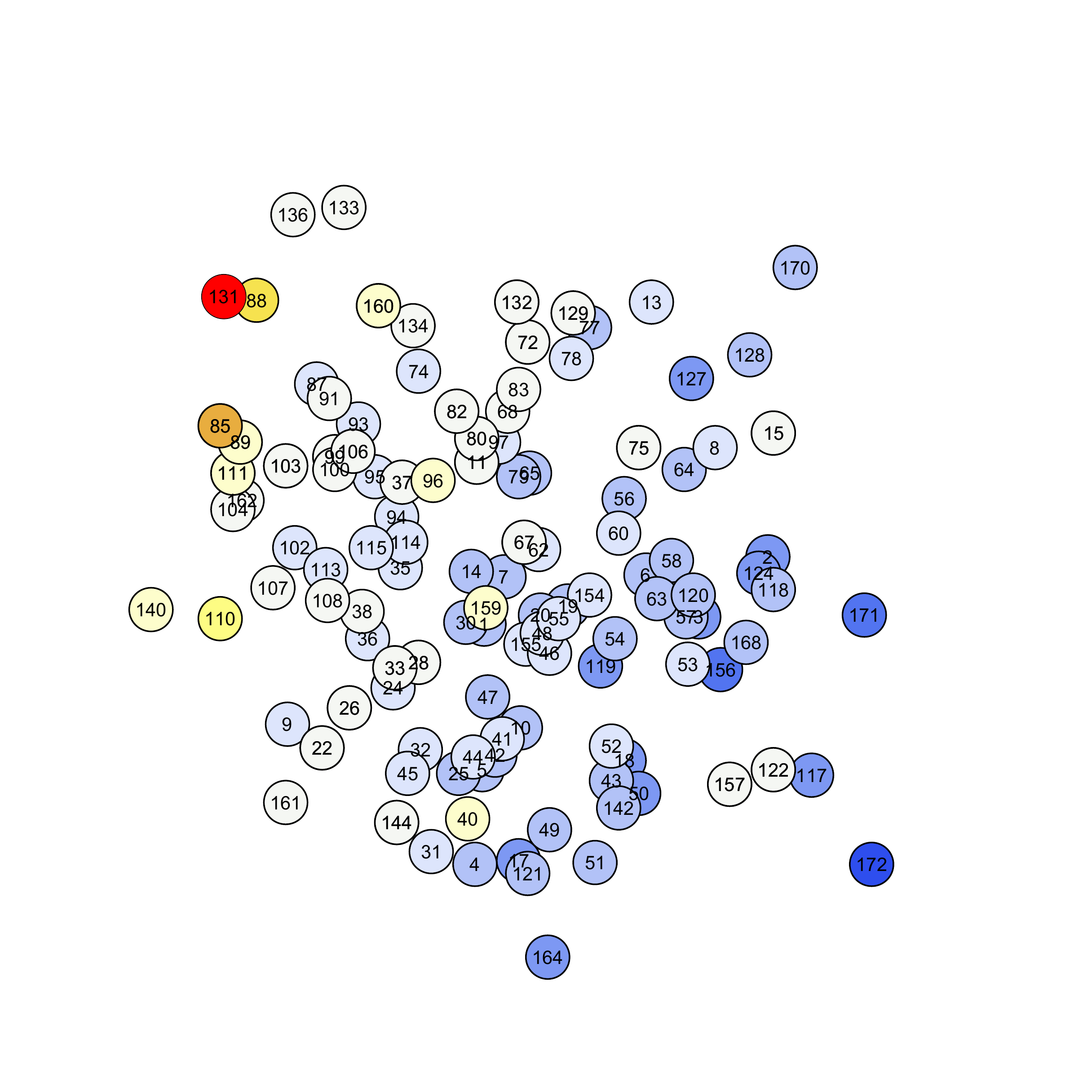}\\
  \end{tabular}
\caption{Firing cells (red) during the frames $190$(top left)-$195$(bottom right) are shown in the figures of layer partition for dataset $I_4$.}\label{firingmove}
\end{figure}

The layer partitions of the estimated graphs by the proposed method for thresholds $\theta = 65.53(I_1), 54.52(I_2), 12.29(I_3), 18.71(I_4)$ are shown in Figure~\ref{LP}, where $\theta$s are determined from estimated distributions of average time delay sum (Figure~\ref{fig:kernel_firing}).The locational direction of layer sequence $V^{\hat{E}}_0, V^{\hat{E}}_1,\cdots$
looks from lower right to upper left for datasets $I_2, I_3$ and $I_4$, which coincides with the move of cells' firing shown in Figure~\ref{firingmove}.

\section{Conclusion and Future Work}

We proposed a method that estimates direct propagation relation between pairs of individuals
from real-valued state sequences of each individual.
Our method calculates time delay sum averaged over all the minimum cost alignments to estimate the direction of state propagation. We believe that our alignment-based method can be applied to analyses of various propagation
by adapting alignment cost calculation to each specific problem.

\begin{acknowledgements}
We would like to thank Prof. Kazuki Horikawa of Tokushima University for giving us a motivation to study the problem treated in this paper. 
We would also like to thank Prof. Tamiki Komatsuzaki for helpful comments to improve this research.
This work was supported by JSPS KAKENHI Grant Number JP18H05413, Japan.

\end{acknowledgements}

%
%

\section*{Declarations}
\small
\begin{description}
   \item[\bf Funding] The authors received support from JSPS KAKENHI Grant Number JP18H05413, Japan. \\
   \item[\bf Conflict of interest] The authors declare that they have no conflict of interest.\\
   \item[\bf Availability of data and material] The data that were used in section~\ref{cellfiring} are available from the corresponding author, A.N., upon reasonable request. \\
   \item[\bf Code availability] The code that was used in this study are available from the corresponding author, A.N. upon reasonable request.
\end{description}

\bibliographystyle{spmpsci}      
\bibliography{ecml2021}   

%
%



\appendix

\section{Calculation of Average Time Delay Sum for the Gap-Based Cost}\label{sec:gap-based-calculation}

In the alignment between two state sequences $s_i$ and $s_j$, either $s_i$ or $s_j$ must not be a null string for the warping-based cost,
but  both $s_i$ and $s_j$ can be null strings  for the gap-based cost.
Thus, the minimum alignment cost $D(t_i,t_j)$ between $s_i[1]\cdots s_i[t_i]$ and  $s_i[1]\cdots s_i[t_j]$ for $t_i=0$ or $t_j=0$
is needed to be calculated, where $s_i[1]\cdots s_i[0]$ represents the null string.
The recursive formula of $D(t_i,t_j)$ for the gap-based cost is the following:
\begin{align*}
 D(t_i,t_j)
  =& \begin{cases}
  0 & \!\!\!\!\!\!\!\!(t_i=t_j=0)\\
  D(t_i,t_j-1)+w(\textvisiblespace,c_j[t_j]) & \!\!\!\!\!\!\!\!(t_i=0, t_j>0)\\
  D(t_i-1,t_j)+w(c_i[t_i],\textvisiblespace) & \!\!\!\!\!\!\!\!(t_i>0, t_j=0)\\
  \min 
  \left\{\!\!\!\begin{array}{l}
  D(t_i-1,t_j)+w(c_i[t_i],\textvisiblespace)\\
  D(t_i,t_j-1)+w(\textvisiblespace,c_j[t_j])\\
  D(t_i-1,t_j-1)+w(c_i[t_i],c_j[t_j])
  \end{array}\!\!\!\right\} & (t_i,t_j>0).
\end{cases}
\end{align*}
The directed graph $G=(V,E)$ whose paths represent the minimum cost alignments
can be constructed as
\begin{align*}
  V=&\{(t_i,t_j)\mid t_i,t_j\in\{0,1,\dots,T\}\}\\
  E=&\begin{aligned}[t]
    & \{((t_i,t_j-1),(t_i,t_j))\mid D(t_i,t_j)=D(t_i,t_j-1)+w(\textvisiblespace,c_j[t_j])\}\\
   & \cup  \{((t_i-1,t_j),(t_i,t_j))\mid D(t_i,t_j)=D(t_i-1,t_j)+w(c_i[t_i],\textvisiblespace)\}\\
   & \cup  \{((t_i-1,t_j-1),(t_i,t_j))\mid  D(t_i,t_j)=D(t_i-1,t_j)+w(c_i[t_i],c_j[t_j])\}.\\
    \end{aligned}
\end{align*}
All the paths from $(0,0)$ to $(T,T)$ on $G$ correspond to the minimum cost alignments.
The number $B(t_i,t_j)$ of paths from $(t_i,t_j)$ to $(T,T)$ can be represented by the same recursive formula as that for the warping-based cost, but the number of the minimum cost alignments between the whole sequences $s_i$ and $s_j$ is $B(0,0)$ instead of $B(1,1)$.
The sum of time delay over matched positions in the minimum cost alignments is calculated by using the following same expression:
\[
\sum_{((t_i-1,t_j-1),(t_i,t_j))\in E^\ast} (t_j-t_i)F(t_i-1,t_j-1)B(t_i,t_j),
\]
where $F(t_i,t_j)$ is the number of paths from $(0,0)$ to $(t_i,t_j)$ in $G$.
The recursive formula of $F(t_i,t_j)$ is 
\begin{align*}
  F(t_i,t_j)
  =&\begin{cases}
1 & \!\!\!\!\!\!\!\!((t_i,t_j)=(0,0))\\
\mathbbm{1}\{((t_i,t_j\!-\!1),(t_i,t_j))\!\in\! E\}F(t_i,t_j\!-\!1) & (t_i\!=\!0, t_j\!>0)\\
\mathbbm{1}\{((t_i-1,t_j),(t_i,t_j))\!\in\! E\}F(t_i\!-\!1,t_j) & (t_i\!>\!0, t_j\!=\!0)\vspace*{1mm}\\
\mathbbm{1}\{((t_i,t_j\!-\!1),(t_i,t_j))\!\in\! E\}F(t_i,t_j\!-\!1) &\\
+\mathbbm{1}\{((t_i\!-\!1,t_j),(t_i,t_j))\!\in\! E\}F(t_i\!-\!1,t_j) &\\
+\mathbbm{1}\{((t_i\!-\!1,t_j\!-\!1),(t_i,t_j))\!\in\! E\}F(t_i\!-\!1,t_j\!-\!1) 
& (t_i,t_j\!>\!0).
\end{cases}
\end{align*}

\begin{figure}[t]
  \begin{center}
    \includegraphics[scale = 0.5]{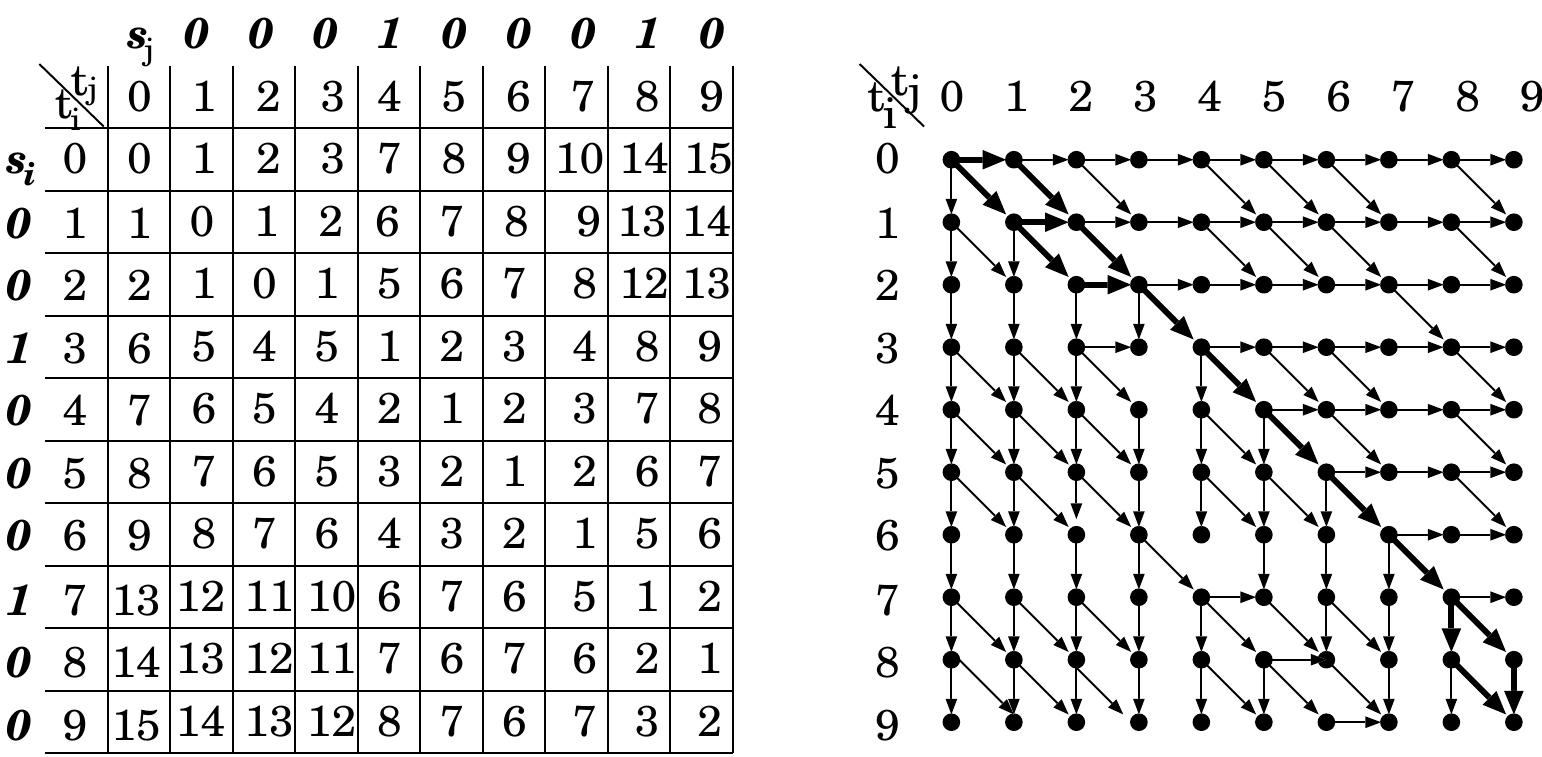}\\
 $D(t_i,t_j)$\hspace*{3.5cm}$G$
    \includegraphics[scale = 0.5]{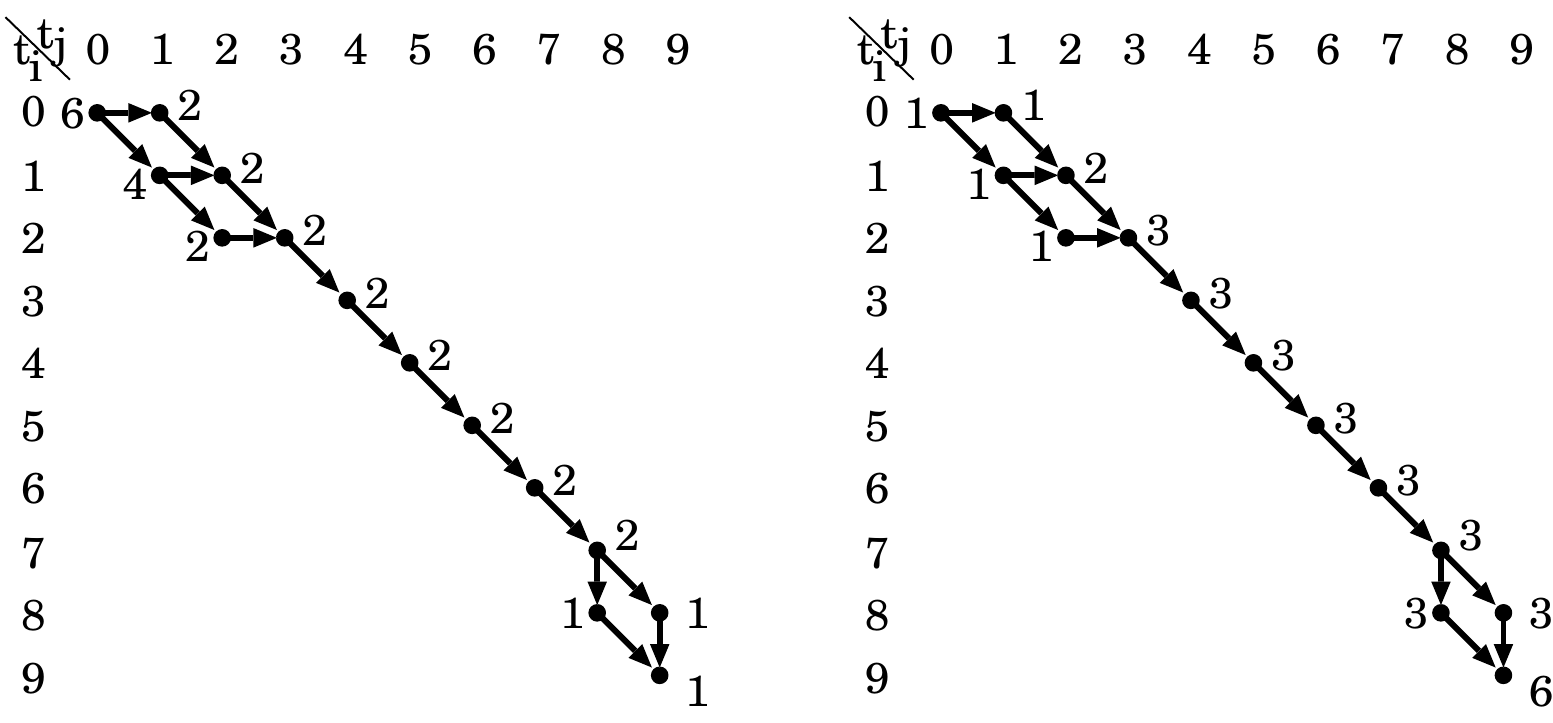}\\
         $B(t_i,t_j)$\hspace*{3cm}$F(t_i,t_j)$
  \end{center}
  \caption{$D$ for strings $s_i$ and $s_j$ with cost function (\ref{costfunction}) setting $\alpha=3$ in Example~\ref{ex:simple} and its corresponding graph $G$, and $B$ and $F$ on the minimum cost paths. The directed edges in the paths from $(0,0)$ to $(9,9)$ on $G$ are bolded. $B(t_i,t_j)$s and $F(t_i,t_j)$s for $(t_i,t_j)$ only in the paths corresponding to the minimum cost alignments are needed to be calculated.}\label{fig:DGBFsimple}
\end{figure}

\begin{example}
  $D$ for strings $s_i$ and $s_j$ with cost function (\ref{costfunction}) seating $\alpha=3$ in Example~\ref{ex:simple}, its corresponding graph $G$,
  the number $B(t_i,t_j)$ of paths from $(t_i,t_j)$ to $(9,9)$ and the number $F(t_i,t_j)$ of paths from $(0,0)$ to $(t_i,t_j)$ on the minimum cost paths in $G$ are shown in Fig.~\ref{fig:DGBFsimple}.
  The minimum cost alignments correspond to the paths from $(0,0)$ to $(9,9)$ in $G$, and the number of those paths is $B(0,0)=F(9,9)=6$.
  From the values in B and F, we can calculate the sum of time delay over matched positions in the minimum cost alignments as
  \begin{align*}
    0\cdot 1\cdot 4 +1\cdot 1\cdot 2 + 0\cdot 1\cdot 2 + 1\cdot 2\cdot 2 +5\times1\cdot 3\cdot 2 &\\
    +1\cdot 3\cdot 1 + 0\cdot 3\cdot 1 &=39.
  \end{align*}
  Thus, the time delay sum averaged over the minimum cost alignments is $39/6=6.5$, which coincides with calculation in Example~\ref{ex:simple}.  
\end{example}

\end{document}